%% file: main.tex
\documentclass[conference]{IEEEtran}
\IEEEoverridecommandlockouts
\usepackage{cite}
\usepackage[colorlinks=true]{hyperref}
\usepackage{amsmath,amssymb,amsfonts}
\usepackage{graphicx}
\usepackage{textcomp}
\usepackage{xcolor}
\usepackage{booktabs}
\usepackage{multirow}
\usepackage{caption}
\usepackage{subfigure}
\usepackage{float}
\usepackage{adjustbox}
\usepackage{bm}
\usepackage[noend]{algpseudocode}
\usepackage{algorithmicx,algorithm}
\usepackage{marvosym}
\hypersetup{urlcolor=black}
\def\BibTeX{{\rm B\kern-.05em{\sc i\kern-.025em b}\kern-.08em
    T\kern-.1667em\lower.7ex\hbox{E}\kern-.125emX}}
\begin{document}

\title{PriSTI: A Conditional Diffusion Framework for Spatiotemporal Imputation \\
}

\author{\IEEEauthorblockN{Mingzhe Liu$^{*1}$, Han Huang$^{*1}$, Hao Feng$^1$, Leilei Sun\textsuperscript{\Letter}$^{1}$, Bowen Du$^{1}$, Yanjie Fu$^{2}$}
\IEEEauthorblockA{$^1$State Key Laboratory of Software Development Environment, Beihang University, Beijing 100191, China\\
$^2$Department of Computer Science, University of Central Florida, FL 32816, USA\\
\{mzliu1997, h-huang, pinghao, leileisun, dubowen\}@buaa.edu.cn, yanjie.fu@ucf.edu
}\thanks{$^*\,$Equal contribution. \Letter $\,$ Corresponding author.}
}

\maketitle

\begin{abstract}
Spatiotemporal data mining plays an important role in air quality monitoring, crowd flow modeling, and climate forecasting. However, the originally collected spatiotemporal data in real-world scenarios is usually incomplete due to sensor failures or transmission loss. Spatiotemporal imputation aims to fill the missing values according to the observed values and the underlying spatiotemporal dependence of them. 
The previous dominant models impute missing values autoregressively and suffer from the problem of error accumulation. 
As emerging powerful generative models, the diffusion probabilistic models can be adopted to impute missing values conditioned by observations and avoid inferring missing values from inaccurate historical imputation. 
However, the construction and utilization of conditional information are inevitable challenges when applying diffusion models to spatiotemporal imputation.
To address above issues, we propose a conditional diffusion framework for spatiotemporal imputation with enhanced prior modeling, named PriSTI. 
Our proposed framework provides a conditional feature extraction module first to extract the coarse yet effective spatiotemporal dependencies from conditional information as the global context prior. Then, a noise estimation module transforms random noise to realistic values, with the spatiotemporal attention weights calculated by the conditional feature, as well as the consideration of geographic relationships. 
PriSTI outperforms existing imputation methods in various missing patterns of different real-world spatiotemporal data, and effectively handles scenarios such as high missing rates and sensor failure.
 The implementation code is available at \url{https://github.com/LMZZML/PriSTI}.
\end{abstract}


\input{1_Introduction}

\input{2_Preliminaries}

\input{3_Methodology}

\input{4_Experiments}

\input{5_Related_Work}

\input{6_Conclusion}

\section*{Acknowledgment}

We thank anonymous reviewers for their helpful comments.
This research is supported by the National Natural Science Foundation of China (62272023).

\bibliographystyle{IEEEtran}
\bibliography{citation}


\end{document}

%% file: 1_Introduction.tex
\section{Introduction}

Spatiotemporal data is a type of data with intrinsic spatial and temporal patterns, which is widely applied in the real world for tasks such as air quality monitoring \cite{cao2018brits, yi2016st}, traffic status forecasting \cite{li2017diffusion, wu2019graph}, weather prediction \cite{bauer2015quiet} and so on. 
However, due to the sensor failures and transmission loss \cite{yi2016st}, the incompleteness in spatiotemporal data is a common problem, characterized by the randomness of missing value's positions and the diversity of missing patterns, which results in incorrect analysis of spatiotemporal patterns and further interference on downstream tasks.  
In recent years, extensive research \cite{cao2018brits, liu2019naomi, cini2021filling} has dived into spatiotemporal imputation, with the goal of exploiting spatiotemporal dependencies from available observed data to impute missing values.

\begin{figure*}[t]
	\center
	\includegraphics[width=0.89\textwidth]{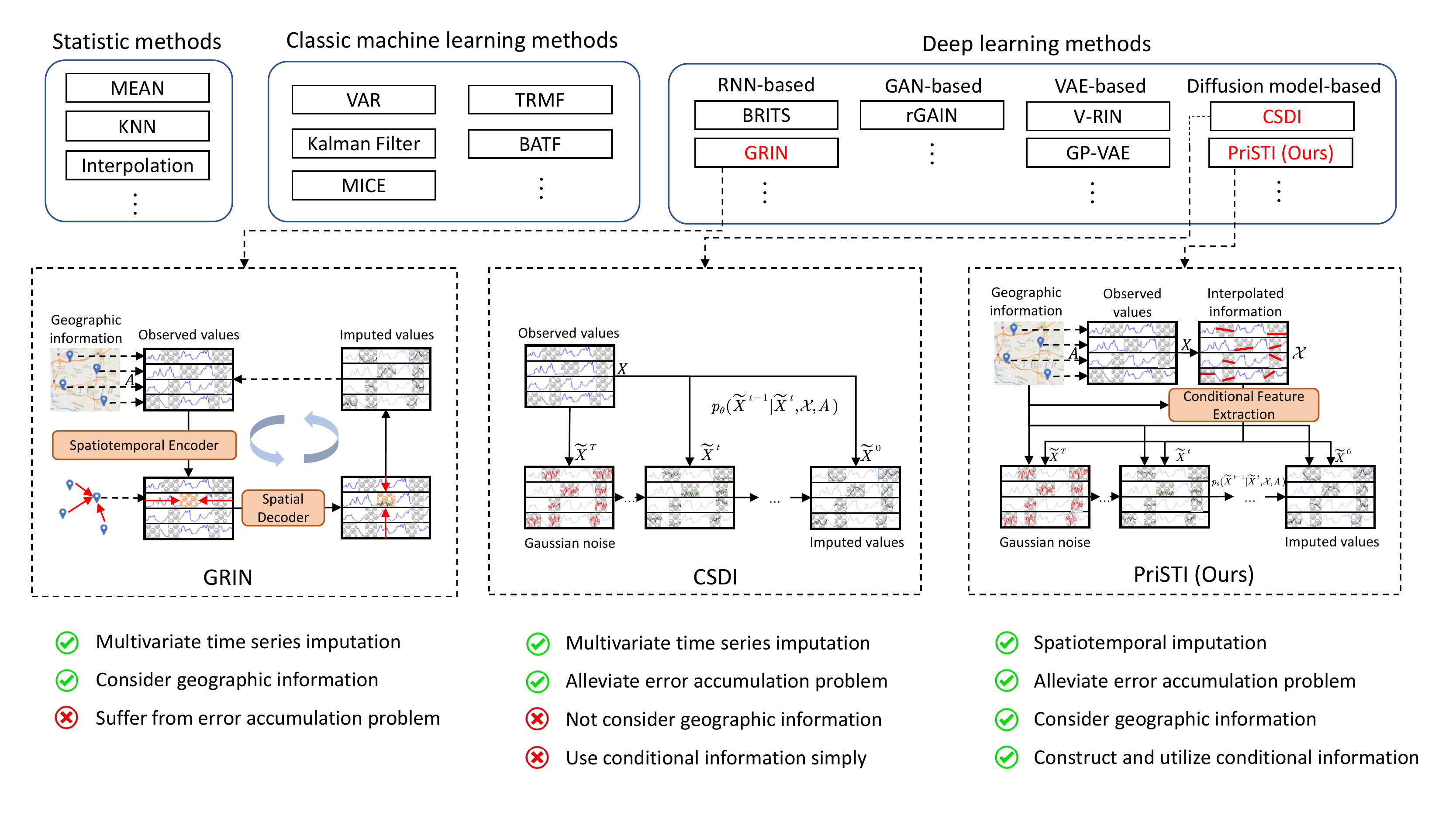}
    \vspace{-5mm}
	\caption{The motivation of our proposed methods. We summarize the existing methods that can be applied to spatiotemporal imputation, and compare our proposed methods with the recent existing methods. The grey shadow represents the missing part, while the rest with blue solid line represents observed values $X$.}\label{fig:motivation}
    \vspace{-3mm}
\end{figure*}

The early studies applied for spatiotemporal imputation usually impute along the temporal or spatial dimension with statistic and classic machine learning methods, including but not limited to autoregressive moving average (ARMA) \cite{ansley1984estimation, harvey1990forecasting}, expectation-maximization algorithm (EM) \cite{shumway1982approach, nelwamondo2007missing}, k-nearest neighbors (KNN) \cite{trevor2009elements, beretta2016nearest}, etc. 
But these methods impute missing values based on strong assumptions such as the temporal smoothness and the similarity between time series, and ignore the complexity of spatiotemporal correlations. 
With the development of deep learning, most effective spatiotemporal imputation methods \cite{cao2018brits, yoon2018gain, cini2021filling} use the recurrent neural network (RNN) as the core to impute missing values by recursively updating their hidden state, capturing the temporal correlation with existing observations.  
Some of them also simply consider the feature correlation \cite{cao2018brits} by the multilayer perceptron (MLP) or spatial similarity between different time series \cite{cini2021filling} by graph neural networks.
However, these approaches inevitably suffer from error accumulation \cite{liu2019naomi}, i.e., inference missing values from inaccurate historical imputation, and only output the deterministic values without reflecting the uncertainty of imputation.

More recently, diffusion probabilistic models (DPM) \cite{sohl2015deep, ho2020denoising, song2020score}, as emerging powerful generative models with impressive performance on various tasks, have been adopted to impute multivariate time series. These methods impute missing values starting from randomly sampled Gaussian noise, and convert the noise to the estimation of missing values \cite{tashiro2021csdi}. 
Since the diffusion models are flexible in terms of neural network architecture, they can circumvent the error accumulation problem from RNN-based methods through utilizing architectures such as attention mechanisms when imputation,
which also have a more stable training process than generative adversarial networks (GAN).
However, when applying diffusion models to imputation problem, the modeling and introducing of the conditional information in diffusion models are the inevitable challenges. For spatiotemporal imputation, the challenges can be specific to the construction and utilization of conditional information with spatiotemporal dependencies.
Tashiro et al. \cite{tashiro2021csdi} only model temporal and feature dependencies by attention mechanism when imputing, without considering spatial similarity such as geographic proximity and time series correlation. 
Moreover, they combine the conditional information (i.e., observed values) and perturbed values directly as the input for models during training, which may lead to inconsistency inside the input spatiotemporal data, increasing the difficulty for the model to learn spatiotemporal dependencies.

To address the above issues, we propose a conditional diffusion framework for SpatioTemporal Imputation with enhanced Prior modeling (PriSTI).
We summarize the existing methods that can be applied to spatiotemporal imputation, and compare the differences between our proposed method and the recent existing methods, as shown in Figure \ref{fig:motivation}.
Since the main challenge of applying diffusion models on spatiotemporal imputation is how to model and utilize the spatiotemporal dependencies in conditional information for the generation of missing values, our proposed method reduce the difficulty of spatiotemporal dependencies learning by extracting conditional feature from observation as a global context prior.
The imputation process of spatiotemporal data with our proposed method is shown in the right of Figure \ref{fig:motivation}, which gradually transform the random noise to imputed missing values by the trained PriSTI.
PriSTI takes observed spatiotemporal data and geographic information as input. During training, the observed values are randomly erased as imputation target through a specific mask strategy.
The incomplete observed data is first interpolated to obtain the enhanced conditional information for diffusion model.
For the construction of conditional information, a conditional feature extraction module is provided to extract the feature with spatiotemporal dependencies from the interpolated information.
Considering the imputation of missing values not only depends on the values of nearby time and similar time series, but also is affected by geographically surrounding sensors, we design the specialized spatiotemporal dependencies learning methods. The proposed method comprehensively aggregates spatiotemporal global features and geographic information to fully exploit the explicit and implicit spatiotemporal relationships in different application scenarios.
For the utilization of conditional information, we design a noise estimation module to mitigate the impact of the added noise on the spatiotemporal dependencies learning. The noise estimation module utilizes the extracted conditional feature, as the global context prior, to calculate the spatiotemporal attention weights, and predict the added Gaussian noise by spatiotemporal dependencies.
PriSTI performs well in spatiotemporal data scenarios with spatial similarity and feature correlation.
For three real-world datasets in the fields of air quality and traffic, our proposed method outperforms existing methods in various missing patterns. Moreover, PriSTI can support downstream tasks through imputation, and effectively handles the case of high missing rates and sensor failure. 

Our contributions are summarized as follows:

\begin{itemize}
    \item We propose PriSTI, a conditional diffusion framework for spatiotemporal imputation, which constructs and utilizes conditional information with spatiotemporal global correlations and geographic relationships.    
   
    \item To reduce the difficulty of learning spatiotemporal dependencies, we design a specialized noise prediction model that extracts conditional features from enhanced observations, calculating the spatiotemporal attention weights using the extracted global context prior.

    \item Our proposed method achieves the best performance on spatiotemporal data in various fields, and effectively handles application scenarios such as high missing rates and sensor failure.
\end{itemize}

The rest of this paper is organized as follows. We first state the definition of the spatiotemporal imputation problem and briefly introduce the background of diffusion models in Section \ref{sec:problem_def}. Then we introduce how the diffusion models are applied to spatiotemporal imputation, as well as the details of our proposed framework in Section \ref{sec:method}. Next, we evaluate the performance of our proposed method in various missing patterns in Section \ref{sec:exp}. Finally, we review the related work for spatiotemporal imputation in Section \ref{sec:related_work} and conclude our work in Section \ref{sec:conclusion}.

%% file: 2_Preliminaries.tex
\section{Preliminaries}\label{sec:problem_def}

In this section, we introduce some key definitions in spatiotemporal imputation, state the problem definition and briefly introduce the diffusion probabilistic models.

\textbf{Spatiotemporal data. }
We formalize spatiotemporal data as a sequence $X_{1:L}=\{X_1, X_2,\cdots, X_L\}\in \mathbb{R}^{N\times L}$ over consecutive time, where $X_l\in\mathbb{R}^N$ is the values observed at time $l$ by $N$ observation nodes, such as air monitoring stations and traffic sensors. Not all observation nodes have observed values at time $l$. We use a binary mask $M_l\in\{0,1\}^N$ to represent the observed mask at time $l$, where $m_l^{i,j}=1$ represents the value is observed while $m_l^{i,j}=0$ represents the value is missing. 
Since there is no ground truth for real missing data in practice, we manually select the imputation target $\widetilde{X}\in \mathbb{R}^{N\times L}$ from available observed data for training and evaluation, and identify them with the binary mask $\widetilde{M}\in \mathbb{R}^{N\times L}$.

\textbf{Adjacency matrix. }
The observation nodes can be formalized as a graph $G=\langle V,E\rangle$, where $V$ is the node set and $E$ is the edge set measuring the pre-defined spatial relationship between nodes, such as geographical distance. 
We denote $A\in \mathbb{R}^{N\times N}$ to represent the geographic information as the adjacency matrix of the graph $G$. In this work, we only consider the setting of static graph, i.e., the geographic information $A$ does not change over time.

\textbf{Problem statement. }
Given the incomplete observed spatiotemporal data $X$ and geographical information $A$, our task of spatiotemporal imputation is to estimate the missing values or corresponding distributions in spatiotemporal data $X_{1:L}$.

\textbf{Diffusion probabilistic models. }
Diffusion probabilistic models \cite{dickstein15, ho2020denoising} are deep generative models that have achieved cutting-edge results in the field of image synthesis \cite{rombach2022high}, audio generation \cite{kong2020diffwave}, etc., which generate samples consistent with the original data distribution by adding noise to the samples and learning the reverse denoising process.  
The diffusion probabilistic model can be formalized as two Markov chain processes of length $T$, named the \textit{diffusion process} and the \textit{reverse process}.
Let $\widetilde{X}^0\sim p_{data}$ where $p_{data}$ is the clean data distribution, and $\widetilde{X}^t$ is the sampled latent variable sequence, where $t=1,\cdots,T$ is the diffusion step. $\widetilde{X}^T\sim \mathcal{N}(0, \bm{I})$ where $\mathcal{N}$ is Gaussian distribution. The diffusion process adds Gaussian noise gradually into $\widetilde{X}^0$ until $\widetilde{X}^0$ is close to $\widetilde{X}^T$, while the reverse process denoises $\widetilde{X}^t$ to recover $\widetilde{X}^0$.
More details about applying the diffusion models on spatiotemporal imputation are introduced in Section \ref{sec:method}.  

\begin{table}[t]
  \centering
  \caption{Important notations and corresponding descriptions.}
  \label{tab:notation}
  \setlength{\tabcolsep}{1mm}
  \resizebox{0.95\columnwidth}{!}{
    \begin{tabular}{c|l}
    \toprule
    Notations & Descriptions\cr
    \midrule    
    $\bm{X}$ & Spatiotemporal data \cr
    $\bm{\widetilde{X}}$ & Manually selected imputation target \cr
    $N$ & The number of the observation nodes \cr
    $L, l$ & Length of the observed time and observed time step \cr
    $T, t$ & Length of the diffusion steps and diffusion step \cr
    $\bm{A}$ & Adjacency matrix of geographic information \cr
    $\mathcal{X}$ & Interpolated conditional information \cr
    $\beta_t, \alpha_t, \hat{\alpha}_t$ & Constant hyperparameters of diffusion model \cr
    $\epsilon_{\theta}$ & Noise prediction model \cr
   
    \bottomrule
    \end{tabular}}
\end{table}

%% file: 3_Methodology.tex
\section{Methodology}\label{sec:method}

\begin{figure}[t]
	\center
	\includegraphics[width=0.9\columnwidth]{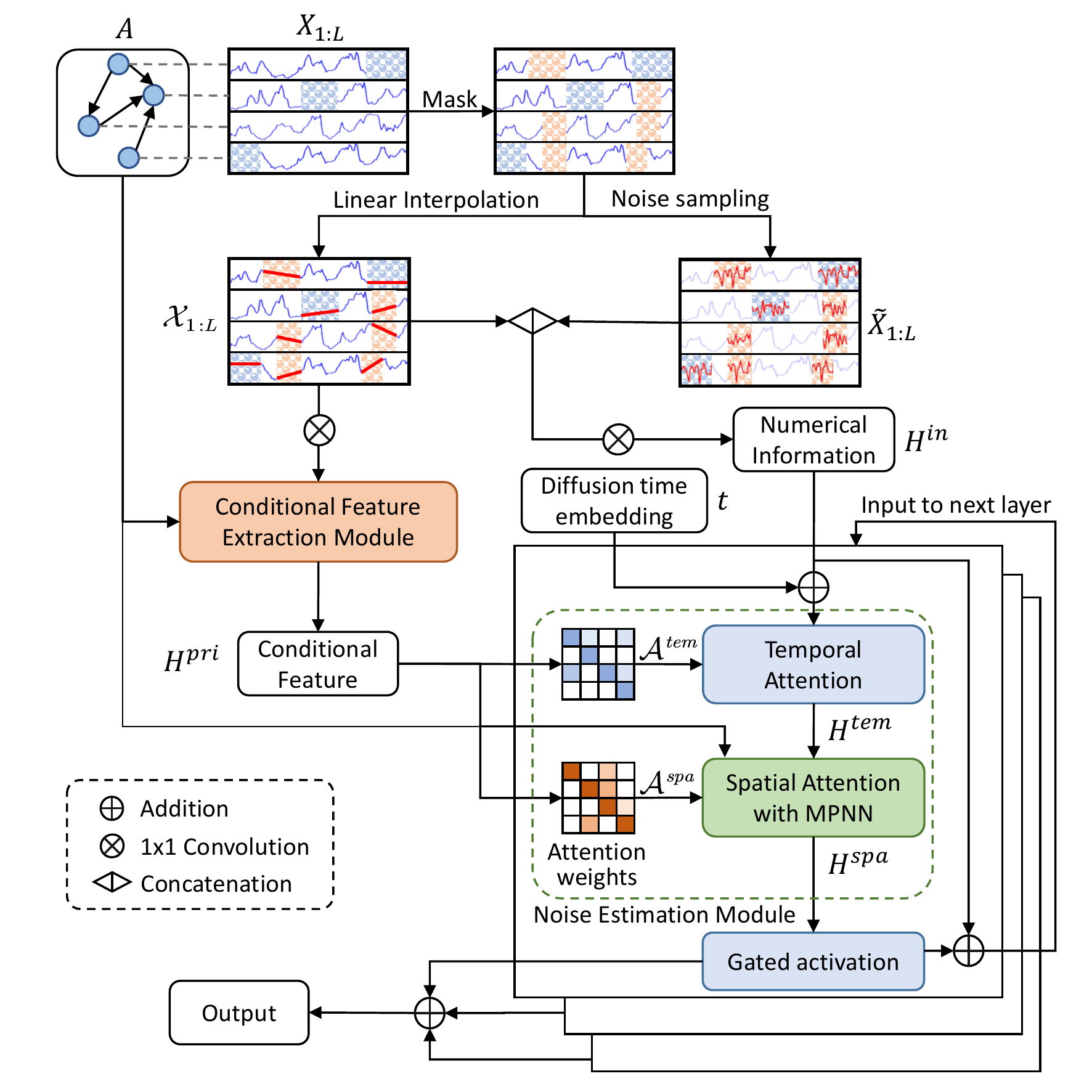}
	\caption{The pipeline of PriSTI. PriSTI takes observed values and geographic information as input. It first interpolates observations and models the global context prior by the conditional feature extraction module, and then utilizes the noise estimation module to predict noise with help of the conditional information.}\label{fig:framework}
\end{figure}

The pipeline of our proposed spatiotemporal imputation framework, PriSTI, is shown in Figure \ref{fig:framework}. PriSTI adopts a conditional diffusion framework to exploit spatiotemporal global correlation and geographic relationship for imputation. To address the challenge about the construction and utilization of conditional information when impute by diffusion model, we design a specialized noise prediction model to enhance and extract the conditional feature. 
In this section, we first introduce how the diffusion models are applied to spatiotemporal imputation, and then introduce the detail architecture of the noise prediction model, which is the key to success of diffusion models.

\subsection{Diffusion Model for Spatiotemporal Imputation}\label{sec:ddpm4imp}

To apply the diffusion models on spatiotemporal imputation, we regard the spatiotemporal imputation problem as a conditional generation task.
The previous studies \cite{tashiro2021csdi} have shown the ability of conditional diffusion probabilistic model for multivariate time series imputation. The spatiotemporal imputation task can be regarded as calculating the conditional probability distribution $q(\widetilde{X}_{1:L}^0|X_{1:L})$, where the imputation of $\widetilde{X}_{1:L}^0$ is conditioned by the observed values $X_{1:L}$.
However, the previous studies impute without considering the spatial relationships, and simply utilize the observed values as conditional information. In this section, we explain how our proposed framework impute spatiotemporal data by the diffusion models.
In the following discussion, we use the superscript $t\in\{0, 1, \cdots, T\}$ to represent the diffusion step, and omit the subscription $1:L$ for conciseness.

As mentioned in Section \ref{sec:problem_def}, the diffusion probabilistic model includes the \textit{diffusion process} and \textit{reverse process}. 
The \textit{diffusion process} for spatiotemporal imputation is irrelavant with conditional information, adding Gaussian noise into original data of the imputation part, which is formalized as:
\begin{equation}
\begin{aligned}
    & q(\widetilde{X}^{1:T}|\widetilde{X}^{0})=\prod_{t=1}^T q(\widetilde{X}^{t}|\widetilde{X}^{t-1}), \\
    & q(\widetilde{X}^{t}|\widetilde{X}^{t-1})=\mathcal{N}(\widetilde{X}^t; \sqrt{1-\beta_t}\widetilde{X}^{t-1}, \beta_t \bm{I}),
\end{aligned}
\end{equation}
where $\beta_t$ is a small constant hyperparameter that controls the variance of the added noise.
The $\widetilde{X}^t$ is sampled by $\widetilde{X}^t=\sqrt{\bar{\alpha}_t}\widetilde{X}^0+\sqrt{1-\bar{\alpha}_t}\epsilon$, where $\alpha_t=1-\beta_t$, $\bar{\alpha}_t=\prod_{i=1}^t\alpha_i$, and $\epsilon$ is the sampled standard Gaussian noise. When $T$ is large enough, $q(\widetilde{X}^T|\widetilde{X}^0)$ is close to standard normal distribution .

The \textit{reverse process} for spatiotemporal imputation gradually convert random noise to missing values with spatiotemporal consistency based on conditional information. In this work, the reverse process is conditioned on the interpolated conditional information $\mathcal{X}$ that enhances the observed values, as well as the geographical information $A$. The reverse process can be formalized as:
\begin{equation}\label{eq:reverse_process}
\begin{aligned}
    & p_{\theta}(\widetilde{X}^{0:T-1}|\widetilde{X}^{T}, \mathcal{X}, A)=\prod_{t=1}^T p_{\theta}(\widetilde{X}^{t-1}|\widetilde{X}^{t}, \mathcal{X}, A), \\
    & p_{\theta}(\widetilde{X}^{t-1}|\widetilde{X}^{t}, \mathcal{X}, A)=\mathcal{N}(\widetilde{X}^{t-1}; \mu_{\theta}(\widetilde{X}^{t}, \mathcal{X}, A, t), \sigma_t^2 \bm{I}).
\end{aligned}
\end{equation}

Ho et al. \cite{ho2020denoising} introduce an effective parameterization of $\mu_{\theta}$ and $\sigma_t^2$. In this work, they can be defined as:
\begin{equation}\label{eq:mu_sigma}
\begin{aligned}
    & \mu_{\theta}(\widetilde{X}^{t}, \mathcal{X}, A, t)=\frac{1}{\sqrt{\bar{\alpha}_t}}\left(\widetilde{X}^{t}-\frac{\beta_t}{\sqrt{1-\bar{\alpha}_t}}\epsilon_{\theta}(\widetilde{X}^{t}, \mathcal{X}, A, t)\right), \\
    & \sigma_t^2=\frac{1-\bar{\alpha}_{t-1}}{1-\bar{\alpha}_t}\beta_t,
\end{aligned}
\end{equation}  
where $\epsilon_{\theta}$ is a neural network parameterized by $\theta$, which takes the noisy sample $\widetilde{X}^{t}$ and conditional information $\mathcal{X}$ and adjacency matrix $A$ as input, predicting the added noise $\epsilon$ on imputation target to restore the original information of the noisy sample.
Therefore, $\epsilon_{\theta}$ is often named \textit{noise prediction model}. The noise prediction model does not limit the network architecture, whose flexibility is benificial for us to design the model suitable for spatiotemporal imputation.

\textbf{Training Process.}
During training, we mask the input observed value $X$ through a random mask strategy to obtain the imputation target $\widetilde{X}^t$, and the remaining observations are used to serve as the conditional information for imputation. Similar to CSDI \cite{tashiro2021csdi}, we provide the mask strategies including point strategy, block strategy and hybrid strategy (see more details in Section \ref{sec:exp_set}). Mask strategies produce different masks for each training sample. 
After obtaining the training imputation target $\widetilde{X}^0$ and the interpolated conditional information $\mathcal{X}$, the training objective of spatiotemporal imputation is:
\begin{equation}\label{eq:loss}
\mathcal{L}(\theta)=\mathbb{E}_{\widetilde{X}^0\sim q(\widetilde{X}^0), \epsilon\sim\mathcal{N}(0,I)}\left\Vert\epsilon-\epsilon_{\theta}(\widetilde{X}^{t}, \mathcal{X}, A, t)\right\Vert^2.
\end{equation}  

Therefore, in each iteration of the training process, we sample the Gaussian noise $\epsilon$, the imputation target $\widetilde{X}^t$ and the diffusion step $t$, and obtain the interpolated conditional information $\mathcal{X}$ based on the remaining observations. 
More details on the training process of our proposed framework is shown in Algorithm \ref{alg:train}.

\begin{algorithm}[t]
	\caption{Training process of PriSTI.}
	\label{alg:train}
	\hspace*{0.02in} {\bf Input:}Incomplete observed data $X$, the adjacency matrix $A$, the number of iteration $N_{it}$, the number of diffusion steps $T$, noise levels sequence $\bar{\alpha}_t$. \\
	\hspace*{0.02in} {\bf Output:}{Optimized noise prediction model $\epsilon_{\theta}$.}  
    \begin{algorithmic}[1]
    \For {$i=1$ \text{to} $N_{it}$}    
        \State $\widetilde{X}^0 \gets \text{Mask}(X)$; 

        \State $\mathcal{X} \gets \text{Interpolate}(\widetilde{X}^0)$;

        \State Sample $t \sim \text{Uniform}(\{1,\cdots,T\})$, $\epsilon\sim\mathcal{N}(0,\textbf{\text{I}})$;


        \State $\widetilde{X}^t \gets \sqrt{\bar{\alpha}_t}\widetilde{X}^0+\sqrt{1-\bar{\alpha}_t}\epsilon$;

        \State Updating the gradient $\nabla_{\theta}\left\Vert\epsilon-\epsilon_{\theta}(\widetilde{X}^{t}, \mathcal{X}, A, t)\right\Vert^2$. 
        
    \EndFor
    
    \end{algorithmic}
\end{algorithm}

\textbf{Imputation Process.}
When using the trained noise prediction model $\epsilon_{\theta}$ for imputation, the observed mask $\widetilde{M}$ of the data is available, so the imputation target $\widetilde{X}$ is the all missing values in the spatiotemporal data, and the interpolated conditional information $\mathcal{X}$ is constructed based on all observed values.  
The model receives $\widetilde{X}^T$ and $\mathcal{X}$ as inputs and generates samples of the imputation results through the process in Equation (\ref{eq:reverse_process}).
The more details on the imputation process of our proposed framework is shown in Algorithm \ref{alg:impute}.

\begin{algorithm}[t]
	\caption{Imputation process with PriSTI.}
	\label{alg:impute}
	\hspace*{0.02in} {\bf Input:}A sample of incomplete observed data $X$, the adjacency matrix $A$, the number of diffusion steps $T$, the optimized noise prediction model $\epsilon_{\theta}$.\\
	\hspace*{0.02in} {\bf Output:}{Missing values of the imputation target $\widetilde{X}^0$.}  
    \begin{algorithmic}[1]
    \State $\mathcal{X} \gets \text{Interpolate}(X)$;
    
    \State Set $\widetilde{X}^T\sim\mathcal{N}(0, \textbf{\text{I}})$;

    \For {$t=T$ \text{to} $1$}
        \State $\mu_{\theta}(\widetilde{X}^{t}, \mathcal{X}, A, t) \gets \frac{1}{\sqrt{\bar{\alpha}_t}}\left(\widetilde{X}^{t}-\frac{\beta_t}{\sqrt{1-\bar{\alpha}_t}}\epsilon_{\theta}(\widetilde{X}^{t}, \mathcal{X}, A, t)\right)$

        \State $\widetilde{X}^{t-1} \gets \mathcal{N}(\mu_{\theta}(\widetilde{X}^{t}, \mathcal{X}, A, t), \sigma_t^2 \bm{I})$ 
    	
    \EndFor
    \end{algorithmic}
\end{algorithm}

Through the above framework, the diffusion model can be applied to spatiotemporal imputation with the conditional information. However, the construction and utilization of conditional information with spatiotemporal dependencies are still challenging. It is necessary to design a specialized noise prediction model $\epsilon_{\theta}$ to reduce the difficulty of learning spatiotemporal dependencies with noisy information, which will be introduced in next section.

\subsection{Design of Noise Prediction Model}

In this section, we illustrate how to design the noise prediction model $\epsilon_{\theta}$ for spatiotemporal imputation.
Specifically, we first interpolate the observed value to obtain the enhanced coarse conditional information. Then, a \textit{conditional feature extraction module} is designed to model the spatiotemporal correlation from the coarse interpolation information. The output of the conditional feature extraction module is utilized in the designed \textit{noise estimation module} to calculate the attention weights, which provides a better global context prior for spatiotemporal dependencies learning.

\subsubsection{Conditional Feature Extraction Module}

\begin{figure}[t]
	\center
	\includegraphics[width=0.9\columnwidth]{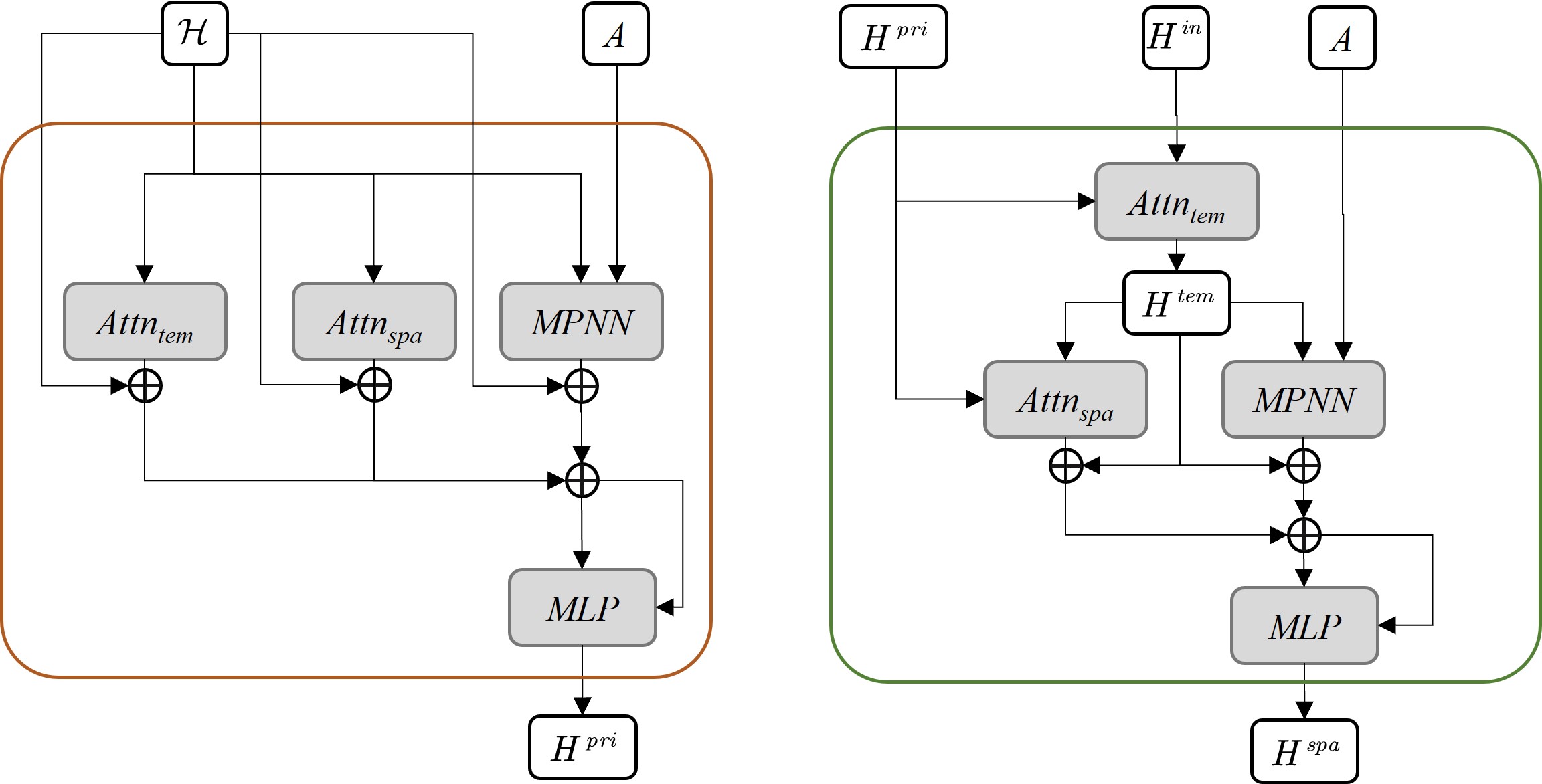}
	\caption{The architecture of the conditional feature extraction module (left) and noise estimation module (right). Both modules utilize the same components, including a temporal attention $\text{Attn}_{tem}$, a spatial attention $\text{Attn}_{spa}$ and a message passing neural network $\text{MPNN}$. The conditional feature extraction module extract spatiotemporal dependencies from interpolated conditional information in a wide network architecture, while the noise estimation module learning spatiotemporal dependencies in a deep network architecture.}\label{fig:STDL}
\end{figure}

The conditional feature extraction module is dedicated to model conditional information when the diffusion model is applied to spatiotemporal imputation.
According to the diffusion model for imputation described above,  $\epsilon_{\theta}$ takes the conditional information $\mathcal{X}$ and the noisy information $\widetilde{X}^t$ as input. The previous studies, such as CSDI \cite{tashiro2021csdi}, regards the observed values as conditional information, and takes the concatenation of conditional information and perturbed values as the input, which are distinguished only by a binary mask.
However, the trend of the time series in imputation target are unstable due to the randomness of the perturbed values, which may cause the noisy sample to have an inconsistent trend with the original time series (such as the $\widetilde{X}_{1:L}^t$ in Figure \ref{fig:framework}), especially when the diffusion step $t$ is close to $T$. 
Although CSDI utilizes two different Transformer layers to capture the temporal and feature dependencies, the mixture of conditional and noisy information increases the learning difficulty of the noise prediction model, which can not be solved by a simple binary mask identifier.

To address the above problem, we first enhance the obvserved values for the conditional feature extraction, expecting the designed model to learn the spatiotemporal dependencies based on this enhanced information.
In particular, inspired by some spatiotemporal forecasting works based on temporal continuity \cite{choi2022graph}, we apply linear interpolation to the time series of each node to initially construct a coarse yet effective interpolated conditional information $\mathcal{X}$ for denoising. 
Intuitively, this interpolation does not introduce randomness to the time series, while also retaining a certain spatiotemporal consistency.
From the test results of linear interpolation on the air quality and traffic speed datasets (see Table \ref{tab:overallmae}), the spatiotemporal information completed by the linear interpolation method is available enough for a coarse conditional information. 
Moreover, the fast computation of linear interpolation satisfies the training requirements of real-time construction under random mask strategies in our framework.

Although linear interpolation solves the completeness of observed information simply and efficiently, it only simply describes the linear uniform change in time, without modeling temporal nonlinear relationship and spatial correlations. 
Therefore, we design a learnable module $\gamma(\cdot)$ to model a conditional feature $H^{pri}$ with spatiotemporal information as a global context prior, named Conditional Feature Extraction Module. 
The module $\gamma(\cdot)$ takes the interpolated conditional information $\mathcal{X}$ and adjacency matrix $A$ as input, extract the spatiotemporal dependencies from $\mathcal{X}$ and output $H^{pri}$ as the global context for the calculation of spatiotemporal attention weights in noise prediction.

In particular, the conditional feature $H^{pri}\in\mathbb{R}^{N\times L\times d}$ is obtained by $H^{pri}=\gamma(\mathcal{H}, A)$, where $\mathcal{H}=\text{Conv}(\mathcal{X})$ and $\text{Conv}(\cdot)$ is $1\times1$ convolution, $\mathcal{H}\in \mathbb{R}^{N\times L\times d}$, and $d$ is the channel size.
The conditional feature extraction module $\gamma(\cdot)$ comprehensively combines the spatiotemporal global correlations and geographic dependency, as shown in the left of Figure \ref{fig:STDL}, which is formalized as:
\begin{equation}\label{eq:gps}
\begin{aligned}
    & H^{pri} = \gamma(\mathcal{H}, A)=\text{MLP}(\varphi_{\text{SA}}(\mathcal{H})+\varphi_{\text{TA}}(\mathcal{H})+\varphi_{\text{MP}}(\mathcal{H}, A)),\\
    & \varphi_{\text{SA}}(\mathcal{H})=\text{Norm}(\text{Attn}_{spa}(\mathcal{H})+\mathcal{H}),\\
    & \varphi_{\text{TA}}(\mathcal{H})=\text{Norm}(\text{Attn}_{tem}(\mathcal{H})+\mathcal{H}),\\
    & \varphi_{\text{MP}}(\mathcal{H}, A)=\text{Norm}(\text{MPNN}(\mathcal{H}, A)+\mathcal{H}),\\
\end{aligned}
\end{equation}
where $\text{Attn}(\cdot)$ represents the global attention, and the subscripts $spa$ and $tem$ represent spatial attention and temporal attention respectively. We use the dot-product multi-head self-attention in Transformer \cite{vaswani2017attention} to implement $\text{Attn}(\cdot)$. And $\text{MPNN}(\cdot)$ represents the spatial message passing neural network, which can be implemented by any graph neural network. We adopt the graph convolution module from Graph Wavenet \cite{wu2019graph}, whose adjacency matrix includes a bidirectional distance-based matrix and an adaptively learnable matrix.

The extracted conditional feature $H^{pri}$ solves the problem of constructing conditional information. It does not contain the added Gaussian noise, and includes temporal dependencies, spatial global correlations and geographic dependencies compared with observed values. To address the remaining challenge of utilizing conditional information, $H^{pri}$ serves as a coarse prior to guide the learning of spatiotemporal dependencies, which is introduced in the next section.

\subsubsection{Noise Estimation Module}

The noise estimation module is dedicated to the utilization of conditional information when the diffusion model is applied to spatiotemporal imputation.
Since the information in noisy sample may have a wide deviation from the real spatiotemporal distribution because of the randomness of the Gaussian noise, it is difficult to learn spatiotemporal dependencies directly from the mixture of conditional and noisy information.
Our proposed noise estimation module captures spatiotemporal global correlations and geographical relationships by a specialized attention mechanism, which reduce the difficulty of spatiotemporal dependencies learning caused by the sampled noise.

Specifically, the inputs of the noise estimation module include two parts: the noisy information $H^{in}=\text{Conv}(\mathcal{X} || \widetilde{X}^t)$ that consists of interpolation information $\mathcal{X}$ and noise sample $\widetilde{X}^t$, and the prior information including the conditional feature $H^{pri}$ and adjacency matrix $A$. 
To comprehensively consider the spatiotemporal global correlation and geographic relationship of the missing data, the temporal features $H^{tem}$ are first learned through a temporal dependency learning module $\gamma_{\mathcal{T}}(\cdot)$, and then the temporal features are aggregated through a spatial dependency learning module $\gamma_{\mathcal{S}}(\cdot)$.
The architecture of the noise estimation module is shown as the right of Figure \ref{fig:STDL}, which is formalized as follows:
\begin{equation}\label{eq:gps_deep}
\begin{aligned}
    & H^{tem}=\gamma_{\mathcal{T}}(H^{in})=\text{Attn}_{tem}(H^{in}), \\
    & H^{spa}=\gamma_{\mathcal{S}}(H^{tem}, A)=\text{MLP}(\varphi_{\text{SA}}(H^{tem})+\varphi_{\text{MP}}(H^{tem}, A)),
\end{aligned}
\end{equation}
where $\text{Attn}_{tem}(\cdot)$, $\varphi_{\text{SA}}(\cdot)$ and $\varphi_{\text{MP}}(\cdot)$ are same as the components in Equation (\ref{eq:gps}), which are used to capture spatiotemporal global attention and geographic similarity, and $H^{tem}, H^{spa}\in \mathbb{R}^{N \times L \times d}$ are the outputs of temporal and spatial dependencies learning modules.

However, in Eq. (\ref{eq:gps_deep}), spatiotemporal dependencies learning is performed on the mixture of conditional noisy information, i.e. $H^{in}$. When the diffusion step $t$ approaches $T$, noise sample $\widetilde{X}^t$ would increase the difficulty of spatiotemporal dependencies learning. To reduce the impact of $\widetilde{X}^t$ while convert it into Gaussian noise, 
we change the input of the attention components $\text{Attn}_{tem}(\cdot)$ and $\text{Attn}_{spa}(\cdot)$, which calculate the attention weights by using the conditional feature $H^{pri}$.  
In particular, take temporal attention $\text{Attn}_{tem}(\cdot)$ as an example, we rewrite the dot-product attention $\text{Attn}_{tem}(Q_{\mathcal{T}},K_{\mathcal{T}},V_{\mathcal{T}})=\text{softmax}(\frac{Q_{\mathcal{T}}K_{\mathcal{T}}^T}{\sqrt{d}})\cdot V_{\mathcal{T}}$ as $\text{Attn}_{tem}(\mathcal{A}_{\mathcal{T}}, V_{\mathcal{T}}) = \mathcal{A}_{\mathcal{T}} \cdot V_{\mathcal{T}}$, where $\mathcal{A}_{\mathcal{T}}=\text{softmax}(\frac{Q_{\mathcal{T}}K_{\mathcal{T}}^T}{\sqrt{d}})$ is the attention weight. 
We calculate the attention weight $\mathcal{A}_{\mathcal{T}}$ by the conditional feature $H^{pri}$, i.e., we set the input $Q_{\mathcal{T}}$, $K_{\mathcal{T}}$ and $V_{\mathcal{T}}$ as:
\begin{equation}\label{eq:cross_att}
Q_{\mathcal{T}}=H^{pri}\cdot W^Q_{\mathcal{T}}, K_{\mathcal{T}}=H^{pri}\cdot W^K_{\mathcal{T}}, V_{\mathcal{T}}=H^{in}\cdot W^V_{\mathcal{T}},
\end{equation}
where $W^Q_{\mathcal{T}}, W^K_{\mathcal{T}}, W^V_{\mathcal{T}}\in\mathbb{R}^{d\times d}$ are learnable projection parameters. 
The spatial attention $\text{Attn}_{spa}(\mathcal{A}_{\mathcal{S}}, V_{\mathcal{S}})$ calculates the attention weight in the same way:
\begin{equation}\label{eq:cross_att_spa}
Q_{\mathcal{S}}=H^{pri}\cdot W^Q_{\mathcal{S}}, K_{\mathcal{S}}=H^{pri}\cdot W^K_{\mathcal{S}}, V_{\mathcal{S}}=H^{tem}\cdot W^V_{\mathcal{S}}.
\end{equation}

The noise estimation module consists of Equation (\ref{eq:gps_deep}) - (\ref{eq:cross_att_spa}), which has the same attention and MPNN components as the conditional feature extraction module with different input and architecture.
The conditional feature extraction module models spatiotemporal dependencies only from the interpolated conditional information $\mathcal{X}$ in a single layer, so it extracts information through a wide network architecture, i.e., directly aggregates the spatiotemporal global correlation and geographic dependency.
Since the noise estimation module needs to convert the noisy sample to standard Gaussian distribution in multiple layers, it learns spatiotemporal dependencies from the noisy samples with help of the conditional feature through a deep network architecture, i.e., extracts the temporal correlation first and aggregates the temporal feature through the spatial global correlation and geographic information.

In addition, when the number of nodes in the spatiotemporal data is large, the computational cost of spatial global attention is high, and the time complexity of its similarity calculation and weighted summation are both $O(N^2d)$. 
Therefore, we map $N$ nodes to $k$ virtual nodes, where $k<N$. 
We rewrite the $K_{\mathcal{S}}$ and $V_{\mathcal{S}}$ in Equation (\ref{eq:cross_att_spa}) when attention is used for spatial dependencies learning as:
\begin{equation}\label{eq:node_samp}
K_{\mathcal{S}}=H^{pri}\cdot P^K_{\mathcal{S}} W^K_{\mathcal{S}} , V_{\mathcal{S}}= H^{tem}\cdot P^V_{\mathcal{S}} W^V_{\mathcal{S}},
\end{equation}
where $P^K_{\mathcal{S}}, P^V_{\mathcal{S}}\in\mathbb{R}^{N\times d}$ is the downsampling parameters. And the time complexity of the modified spatial attention is reduced to $O(Nkd)$.

\subsubsection{Auxiliary Information and Output}
We add auxiliary information $U=\text{MLP}(U_{tem}, U_{spa})$ to both the conditional feature extraction module and the noise estimation module to help the imputation, where $U_{tem}$ is the sine-cosine temporal encoding \cite{vaswani2017attention}, and $U_{spa}$ is learnable node embedding.  
We expand and concatenate $U_{tem}\in\mathbb{R}^{L\times 128}$ and $U_{spa}\in\mathbb{R}^{N\times 16}$, and obtain auxiliary information $U\in\mathbb{R}^{N\times L\times d}$ that can be input to the model through an MLP layer.

The noise estimation module stacks multiple layers, and the output $H^{spa}$ of each layer is divided into residual connection and skip connection after a gated activation unit. The residual connection is used as the input of the next layer, and the skip connections of each layer are added and through two layers of $1\times 1$ convolution to obtain the output of the noise prediction model $\epsilon_{\theta}$. The output only retains the value of imputation target, and the loss is calculated by Equation (\ref{eq:loss}).

%% file: 4_Experiments.tex
\section{Experiments}\label{sec:exp}

In this section, we first introduce the dataset, baselines, evaluation metrics and settings of our experiment. Then, we evaluate our proposed framework PriSTI with a large amount of experiments for spatiotemporal imputation to answer the following research questions:
\begin{itemize}
    \item \textbf{RQ1}: Can PriSTI provide superior imputation performance in various missing patterns compared to several state-of-the-art baselines?
    \item \textbf{RQ2}: How is the imputation performance for PriSTI for different missing rate of spatiotemporal data?
    \item \textbf{RQ3}: Does PriSTI benefit from the construction and utilization of the conditional information?
    \item \textbf{RQ4}: Does PriSTI extract the temporal and spatial dependencies from the observed spatiotemporal data?
    \item \textbf{RQ5}: Can PriSTI impute the time series for the unobserved sensors only based on the geographic location?
\end{itemize}

\subsection{Dataset}

We conduct experiments on three real-world datasets: an air quality dataset AQI-36, and two traffic speed datasets METR-LA and PEMS-BAY. 
AQI-36 \cite{yi2016st} contains hourly sampled PM2.5 observations from 36 stations in Beijing, covering a total of 12 months. METR-LA \cite{li2017diffusion} contains traffic speed collected by 207 sensors in the highway of Los Angeles County \cite{jagadish2014big} in 4 months, and PEMS-BAY \cite{li2017diffusion} contains traffic speed collected by 325 sensors on highways in the San Francisco Bay Area in 6 months. Both traffic datasets are sampled every 5 minutes.
For the geographic information, the adjacency matrix is obtained based on the geographic distances between monitoring stations or sensors followed the previous works \cite{li2017diffusion}. We build the adjacency matrix for the three datasets using thresholded Gaussian kernel \cite{shuman2013emerging}.

\subsection{Baselines}

To evaluate the performance of our proposed method, we compare with classic models and state-of-the-art methods for spatiotemporal imputation. The baselines include statistic methods (MEAN, DA, KNN, Lin-ITP), classic machine learning methods (MICE, VAR, Kalman), low-rank matrix factorization methods (TRMF, BATF), deep autoregressive methods (BRITS, GRIN) and deep generative methods (V-RIN, GP-VAE, rGAIN, CSDI).
We briefly introduce the baseline methods as follows:

    (1)\textbf{MEAN}: directly use the historical average value of each node to impute.
    (2)\textbf{DA}: impute missing values with the daily average of corresponding time steps.
    (3)\textbf{KNN}: calculate the average value of nearby nodes based on geographic distance to impute. 
    (4)\textbf{Lin-ITP}: linear interpolation of the time series for each node, as implemented by torchcde\footnote{https://github.com/patrick-kidger/torchcde}.
    (5)\textbf{KF}: use Kalman Filter to impute the time series for each node, as implemented by filterpy\footnote{https://github.com/rlabbe/filterpy}.
    (6)\textbf{MICE} \cite{white2011multiple}: multiple imputation method by chain equations; 
    (7)\textbf{VAR}: vector autoregressive single-step predictor. 
    (8)\textbf{TRMF} \cite{yu2016temporal}: a temporal regularized matrix factorization method.  
    (9)\textbf{BATF} \cite{chen2019missing}: a Bayesian augmented tensor factorization model, which incorporates the generic forms of spatiotemporal domain knowledge. We implement TRMF and BATF using the code in the Transdim\footnote{https://github.com/xinychen/transdim} repository. The rank is set to 10, 40 and 50 on AQI-36, METR-LA and PEMS-BAY, respectively.
    (10)\textbf{V-RIN} \cite{mulyadi2021uncertainty}: a method to improve deterministic imputation using the quantified uncertainty of VAE, whose probability imputation result is provided by the quantified uncertainty.  
    (11)\textbf{GP-VAE} \cite{fortuin2020gp}: a method for time series probabilistic imputation by combining VAE with Gaussian process. 
    (12)\textbf{rGAIN}: GAIN \cite{yoon2018gain} with a bidirectional recurrent encoder-decoder, which is a GAN-based method. 
    (13)\textbf{BRITS} \cite{cao2018brits}: a multivariate time series imputation method based on bidirectional RNN. 
    (14)\textbf{GRIN} \cite{cini2021filling}: a bidirectional GRU based method with graph neural network for multivariate time series imputation. 
    (15)\textbf{CSDI} \cite{tashiro2021csdi}: a probability imputation method based on conditional diffusion probability model, which treats different nodes as multiple features of the time series, and using Transformer to capture feature dependencies.

In the experiment, the baselines MEAN, KNN, MICE, VAR, rGAIN, BRITS and GRIN are implemented by the code\footnote{https://github.com/Graph-Machine-Learning-Group/grin} provided by the authors of GRIN \cite{cini2021filling}. 
We reproduced these baselines, and the results are consistent with their claims, so we retained the results claimed in GRIN for the above baselines.
The implementation details of the remaining baselines have been introduced as above.

\subsection{Evaluation metrics}
We apply three evaluation metrics to measure the performance of spatiotemporal imputation: Mean Absolute Error (MAE), Mean Squared Error (MSE) and Continuous Ranked Probability Score (CRPS) \cite{matheson1976scoring}. 
MAE and MSE reflect the absolute error between the imputation values and the ground truth, 
and CRPS evaluates the compatibility of the estimated probability distribution with the observed value.
We introduce the calculation details of CRPS as follows.
For a missing value $x$ whose estimated probability distribution is $D$, CRPS measures the compatibility of $D$ and $x$, which can be defined as the integral of the quantile loss $\Lambda_{\alpha}$:
\begin{equation}
\begin{aligned}
    \text{CRPS}(D^{-1},x) & =\int^1_0 2\Lambda_{\alpha}(D^{-1}(\alpha), x)d\alpha,\\
    \Lambda_{\alpha}(D^{-1}(\alpha), x) & =(\alpha-\mathbb{I}_{x<D^{-1}(\alpha)})(x-D^{-1}(\alpha)),
\end{aligned}
\end{equation}
where $\alpha\in[0,1]$ is the quantile levels, $D^{-1}(\alpha)$ is the $\alpha$-quantile of distribution $D$, $\mathbb{I}$ is the indicator function.
Since our distribution of missing values is approximated by generating 100 samples, we compute quantile losses for discretized quantile levels with 0.05 ticks following \cite{tashiro2021csdi} as:
\begin{equation}
    \text{CRPS}(D^{-1},x) \simeq \sum_{i=1}^{19}2\Lambda_{i\times 0.05}(D^{-1}(i\times 0.05), x)/19.
\end{equation}
We compute CRPS for each estimated missing value and use the average as the evaluation metric, which is formalized as:
\begin{equation}
    \text{CRPS}(D, \widetilde{X})=\frac{\sum_{\tilde{x}\in\widetilde{X}}\text{CRPS}(D^{-1},\tilde{x})}{|\widetilde{X}|}.
\end{equation}

\subsection{Experimental settings}\label{sec:exp_set}

\textbf{Dataset.}
We divide training/validation/test set following the settings of previous work \cite{yi2016st, cini2021filling}. 
For AQI-36, we select Mar., Jun., Sep., and Dec. as the test set, the last 10\% of the data in Feb., May, Aug., and Nov. as the validation set, and the remaining data as the training set. 
For METR-LA and PEMS-BAY, we split the training/validation/test set by $70\%/10\%/20\%$. 

\begin{figure}[h]
	\center
	\includegraphics[width=0.8\columnwidth]{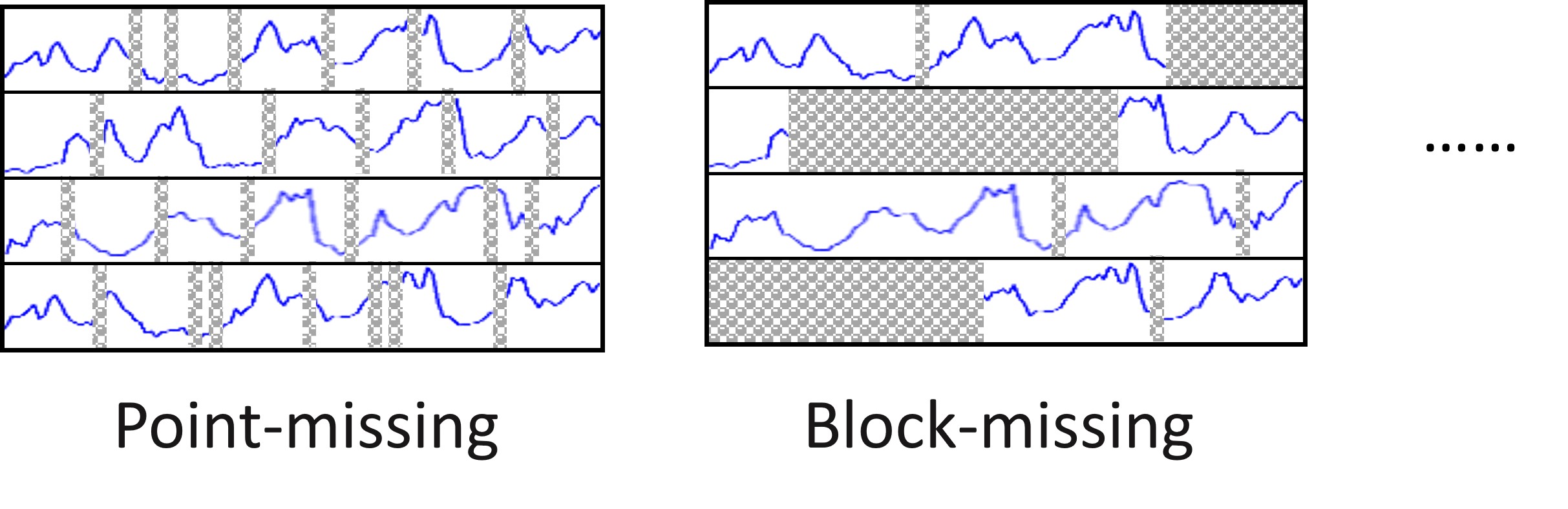}
	\caption{The illustration of some missing patterns.}\label{fig:bp-missing}
\end{figure}

\textbf{Imputation target.}
For air quality dataset AQI-36, we adapt the same evaluation strategy as the previous work provided by \cite{yi2016st}, which simulates the distribution of real missing data. 
For the traffic datasets METR-LA and PEMS-BAY, we use the artificially injected missing strategy provided by \cite{cini2021filling} for evaluation, as shown in Figure \ref{fig:bp-missing}, which includes two missing patterns: 
(1) \textbf{Block missing}: based on randomly masking 5\% of the observed data, mask observations ranging from 1 to 4 hours for each sensor with 0.15\% probability; 
(2) \textbf{Point missing}: randomly mask 25\% of observations. 
The missing rate of each dataset under different missing patterns has been marked in Table \ref{tab:overallmae}. It is worth noting that in addition to the manually injected faults, each dataset also has original missing data (13.24\% in AQI-36, 8.10\% in METR-LA and 0.02\% in PEMS-BAY). All evaluations are performed only on the manually masked parts of the test set.

\textbf{Training strategies.}
As mentioned in Section \ref{sec:ddpm4imp}, on the premise of known missing patterns in test data, we provide three mask strategies.
The details of these mask strategies are described as follows:
\begin{itemize}
    \item Point strategy: take a random value $m$ between [0, 100], randomly select $m$\% of the data from $X$ as the imputation target $\widetilde{X}$, and the remaining unselected data is regarded as observed values in training process.
    \item Block strategy: For each node, a sequence with a length in the range $[L/2, L]$ is selected as the imputation target with a probability from 0 to 15\%. In addition, 5\% of the observed values are randomly selected and added to the imputation target.
    \item Hybrid strategy: For each training sample $X$, it has a 50\% probability to be masked by the point strategy, and the other 50\% probability to be masked by the block strategy or a historical missing pattern i.e., the missing patterns of other samples in the training set. 
\end{itemize}

We utilize different mask strategies for various missing patterns and datasets to make the model simulate the corresponding missing patterns as much as possible during training. Since AQI-36 has much original missing in the training set, which is fewer in traffic datasets, during the training process of PriSTI, we adopt hybrid strategy with historical missing pattern on AQI-36, hybrid strategy with block strategy on block-missing of traffic datasets, and point strategy on point-missing of traffic datasets.

\textbf{Hyperparameters of PriSTI.} For the hyperparameters of PriSTI, the batch size is 16. The learning rate is decayed to 0.0001 at 75\% of the total epochs, and decayed to 0.00001 at 90\% of the total epochs. The hyperparameter for diffusion model include a minimum noise level $\beta_1$ and a maximum noise level $\beta_T$. We adopted the quadratic schedule for other noise levels following \cite{tashiro2021csdi}, which is formalized as:
\begin{equation}
    \beta_t=\left(\frac{T-t}{T-1}\sqrt{\beta_1}+\frac{t-1}{T-1}\sqrt{\beta_T}\right)^2.
\end{equation}

The diffusion time embedding and temporal encoding are implemented by the sine and cosine embeddings followed previous works \cite{kong2020diffwave,tashiro2021csdi}. 
We summary the hyperparameters of PriSTI in Table \ref{tab:exp_setting}. All the experiments are run for 5 times.

\begin{table}[t]
  \centering
  \caption{The hyperparameters of PriSTI for all datasets.}
  \label{tab:exp_setting}
  \setlength{\tabcolsep}{1mm}
  \resizebox{0.9\columnwidth}{!}{
    \begin{tabular}{cccc}
    \toprule
    Description & AQI-36 & METR-LA& PEMS-BAY\cr
    \midrule
    Batch size  & 16 & 16 & 16 \cr
    Time length $L$ & 36 & 24 & 24 \cr
    Epochs & 200 & 300 & 300 \cr
    Learning rate & 0.001 & 0.001 & 0.001 \cr
    Layers of noise estimation & 4 & 4 & 4 \cr
    Channel size $d$ & 64 & 64 & 64 \cr
    Number of attention heads & 8 & 8 & 8 \cr
    Minimum noise level $\beta_1$ & 0.0001 & 0.0001 & 0.0001 \cr
    Maximum noise level $\beta_T$ & 0.2 & 0.2 & 0.2 \cr
    Diffusion steps $T$ & 100 & 50 & 50 \cr
    Number of virtual nodes $k$ & 16 & 64 & 64 \cr
    \bottomrule
    \end{tabular}}
\end{table}

\begin{table*}[ht]
  \centering
  \caption{The results of MAE and MSE for spatiotemporal imputation.}
  \label{tab:overallmae}
  \resizebox{0.95\textwidth}{!}{  
    \setlength{\tabcolsep}{1mm}{
    \renewcommand{\arraystretch}{1}
    \begin{tabular}{ccccccccccc}
    \toprule
    \multirow{3}{*}{Method}&
    \multicolumn{2}{c}{AQI-36}& \multicolumn{4}{c}{METR-LA}& \multicolumn{4}{c}{PEMS-BAY}\cr
    \cmidrule(lr){2-3} \cmidrule(lr){4-7} \cmidrule(lr){8-11}
    & \multicolumn{2}{c}{Simulated failure (24.6\%)}& \multicolumn{2}{c}{Block-missing (16.6\%)}& \multicolumn{2}{c}{Point-missing (31.1\%)}& \multicolumn{2}{c}{Block-missing (9.2\%)}& \multicolumn{2}{c}{Point-missing (25.0\%)}\cr
    \cmidrule(lr){2-3} \cmidrule(lr){4-5} \cmidrule(lr){6-7} \cmidrule(lr){8-9} \cmidrule(lr){10-11}
     & MAE & MSE & MAE & MSE & MAE & MSE & MAE & MSE & MAE & MSE \cr
    \midrule
    Mean & 53.48$\pm$0.00 & 4578.08$\pm$0.00 & 7.48$\pm$0.00 & 139.54$\pm$0.00 & 7.56$\pm$0.00 & 142.22$\pm$0.00 & 5.46$\pm$0.00 & 87.56$\pm$0.00 & 5.42$\pm$0.00 & 86.59$\pm$0.00 \cr
    DA & 50.51$\pm$0.00 & 4416.10$\pm$0.00 & 14.53$\pm$0.00 & 445.08$\pm$0.00 & 14.57$\pm$0.00 & 448.66$\pm$0.00 & 3.30$\pm$0.00 & 43.76$\pm$0.00 & 3.35$\pm$0.00 & 44.50$\pm$0.00  \cr
    KNN & 30.21$\pm$0.00 & 2892.31$\pm$0.00 & 7.79$\pm$0.00 & 124.61$\pm$0.00 & 7.88$\pm$0.00 & 129.29$\pm$0.00 & 4.30$\pm$0.00 & 49.90$\pm$0.00 & 4.30$\pm$0.00 & 49.80$\pm$0.00 \cr
    Lin-ITP & 14.46$\pm$0.00 & 673.92$\pm$0.00 & 3.26$\pm$0.00 & 33.76$\pm$0.00 & 2.43$\pm$0.00 & 14.75$\pm$0.00 & 1.54$\pm$0.00 & 14.14$\pm$0.00 & 0.76$\pm$0.00 & 1.74$\pm$0.00 \cr
    \midrule
    KF & 54.09$\pm$0.00 & 4942.26$\pm$0.00 & 16.75$\pm$0.00 & 534.69$\pm$0.00 & 16.66$\pm$0.00 & 529.96$\pm$0.00 & 5.64$\pm$0.00 & 93.19$\pm$0.00 & 5.68$\pm$0.00 & 93.32$\pm$0.00 \cr
    MICE & 30.37$\pm$0.09 & 2594.06$\pm$7.17 & 4.22$\pm$0.05 & 51.07$\pm$1.25 & 4.42$\pm$0.07 & 55.07$\pm$1.46 & 2.94$\pm$0.02 & 28.28$\pm$0.37 & 3.09$\pm$0.02 & 31.43$\pm$0.41 \cr
    VAR & 15.64$\pm$0.08 & 833.46$\pm$13.85 & 3.11$\pm$0.08 & 28.00$\pm$0.76 & 2.69$\pm$0.00 & 21.10$\pm$0.02 & 2.09$\pm$0.10 & 16.06$\pm$0.73 & 1.30$\pm$0.00 & 6.52$\pm$0.01 \cr
    TRMF & 15.46$\pm$0.06 & 1379.05$\pm$34.83 & 2.96$\pm$0.00 & 22.65$\pm$0.13 & 2.86$\pm$0.00 & 20.39$\pm$0.02 & 1.95$\pm$0.01 & 11.21$\pm$0.06 & 1.85$\pm$0.00 & 10.03$\pm$0.00 \cr
    BATF & 15.21$\pm$0.27 & 662.87$\pm$29.55 & 3.56$\pm$0.01 & 35.39$\pm$0.03 & 3.58$\pm$0.01 & 36.05$\pm$0.02 & 2.05$\pm$0.00 & 14.48$\pm$0.01 & 2.05$\pm$0.00 & 14.90$\pm$0.06  \cr
    \midrule
    V-RIN & 10.00$\pm$0.10 & 838.05$\pm$24.74 & 6.84$\pm$0.17 & 150.08$\pm$6.13 & 3.96$\pm$0.08 & 49.98$\pm$1.30 & 2.49$\pm$0.04 & 36.12$\pm$0.66 & 1.21$\pm$0.03 & 6.08$\pm$0.29  \cr
    GP-VAE & 25.71$\pm$0.30 & 2589.53$\pm$59.14 & 6.55$\pm$0.09 & 122.33$\pm$2.05 & 6.57$\pm$0.10 & 127.26$\pm$3.97 & 2.86$\pm$0.15 & 26.80$\pm$2.10 & 3.41$\pm$0.23 & 38.95$\pm$4.16 \cr
    rGAIN & 15.37$\pm$0.26 & 641.92$\pm$33.89 & 2.90$\pm$0.01 & 21.67$\pm$0.15 & 2.83$\pm$0.01 & 20.03$\pm$0.09 & 2.18$\pm$0.01 & 13.96$\pm$0.20 & 1.88$\pm$0.02 & 10.37$\pm$0.20 \cr
    BRITS & 14.50$\pm$0.35 & 622.36$\pm$65.16 & 2.34$\pm$0.01 & 17.00$\pm$0.14 & 2.34$\pm$0.00 & 16.46$\pm$0.05 & 1.70$\pm$0.01 & 10.50$\pm$0.07 & 1.47$\pm$0.00 & 7.94$\pm$0.03 \cr
    GRIN & 12.08$\pm$0.47 & 523.14$\pm$57.17 & 2.03$\pm$0.00 & 13.26$\pm$0.05 & 1.91$\pm$0.00 & 10.41$\pm$0.03 & 1.14$\pm$0.01 & 6.60$\pm$0.10 & 0.67$\pm$0.00 & 1.55$\pm$0.01 \cr
    CSDI & 9.51$\pm$0.10 & 352.46$\pm$7.50 & 1.98$\pm$0.00 & 12.62$\pm$0.60 & 1.79$\pm$0.00 & 8.96$\pm$0.08 & 0.86$\pm$0.00 & 4.39$\pm$0.02 & 0.57$\pm$0.00 & 1.12$\pm$0.03 \cr
    \midrule
    PriSTI & \textbf{9.03$\pm$0.07} & \textbf{310.39$\pm$7.03} & \textbf{1.86$\pm$0.00} & \textbf{10.70$\pm$0.02} & \textbf{1.72$\pm$0.00} & \textbf{8.24$\pm$0.05} & \textbf{0.78$\pm$0.00} & \textbf{3.31$\pm$0.01} & \textbf{0.55$\pm$0.00} & \textbf{1.03$\pm$0.00} \cr
    \bottomrule
    \end{tabular}}}
\end{table*}

\begin{table}[t]
  \centering
  \caption{The results of CRPS for spatiotemporal imputation.}
  \label{tab:pro_est}
  \renewcommand{\arraystretch}{1}
  \resizebox{0.9\columnwidth}{!}{
    \begin{tabular}{cccccc}
    \toprule
    \multirow{2}{*}{Method}&
    AQI-36& \multicolumn{2}{c}{METR-LA}& \multicolumn{2}{c}{PEMS-BAY}\cr
   \cmidrule(lr){2-2} \cmidrule(lr){3-4} \cmidrule(lr){5-6}
    & {SF}& {Block}& {Point}& {Block}& {Point}\cr
    \midrule
    V-RIN & 0.3154 & 0.1283 & 0.0781 & 0.0394 & 0.0191  \cr
    GP-VAE & 0.3377 & 0.1118 & 0.0977 & 0.0436 & 0.0568  \cr
    CSDI & 0.1056 & 0.0260 & 0.0235 & 0.0127 & 0.0067  \cr
    \midrule
    PriSTI & \textbf{0.0997}  & \textbf{0.0244}  & \textbf{0.0227}  & \textbf{0.0093}  & \textbf{0.0064}  \cr
    \bottomrule
    \end{tabular}}
\end{table}

\begin{table}[t]
  \centering
  \caption{The prediction on AQI-36 after imputation.}
  \label{tab:prediction}
  \renewcommand{\arraystretch}{1}
  \resizebox{0.9\columnwidth}{!}{
    \begin{tabular}{cccccc}
    \toprule
    Metric & Ori. & BRITS & GRIN & CSDI & PriSTI \cr
    \midrule
    MAE & 36.97 & 34.61 & 33.77 & 30.20 & \textbf{29.34}  \cr
    RMSE &60.37 & 56.66 & 54.06 & 46.98 & \textbf{45.08}  \cr
    \bottomrule
    \end{tabular}}
\end{table}

\subsection{Results}

\subsubsection{Overall Performance (RQ1)}

We first evaluate the spatiotemporal imputation performance of PriSTI compared with other baselines. 
Since not all methods can provide the probability distribution of missing values, i.e. evaluated by the CRPS metric, we show the deterministic imputation result evaluated by MAE and MSE in Table \ref{tab:overallmae}, and select V-RIN, GP-VAE, CSDI and PriSTI to be evaluated by CRPS, which is shown in Table \ref{tab:pro_est}. 
The probability distribution of missing values of these four methods are simulated by generating 100 samples, while their deterministic imputation result is the median of all generated samples.
Since CRPS fluctuates less across 5 times experiments (the standard error is less than 0.001 for CSDI and PriSTI), we only show the mean of 5 times experiments in Table \ref{tab:pro_est}. 

It can be seen from Table \ref{tab:overallmae} and Table \ref{tab:pro_est} that our proposed method outperforms other baselines on various missing patterns in different datasets. We summarize our findings as follows:
(1) The statistic methods and classic machine learning methods performs poor on all the datasets. These methods impute missing values based on assumptions such as stability or seasonality of time series, which can not cover the complex temporal and spatial correlations in real-world datasets. 
The matrix factorization methods also perform not well due to the low-rank assumption of data.
(2) Among deep learning methods, GRIN, as an autoregressive state-of-the-art multivariate time series imputation method, performs better than other RNN-based methods (rGAIN and BRITS) due to the extraction of spatial correlations. However, the performance of GRIN still has a gap compared to the diffusion model-based methods (CSDI), which may be caused by the inherent defect of error accumulation in autoregressive models.
(3) For deep generative models, the VAE-based methods (V-RIN and GP-VAE) can not outperform CSDI and PriSTI. Our proposed method PriSTI outperforms CSDI in various missing patterns of every datasets, which indicates that our design of conditional information construction and spatiotemporal correlation can improve the performance of diffusion model for imputation task. 
In addition, for the traffic datasets, we find that our method has a more obvious improvement than CSDI in the block-missing pattern compared with point-missing, which indicates that the interpolated information may provide more effective conditional information than observed values especially when the missing is continuous at time.

In addition, we select the methods of the top 4 performance rankings (i.e., the average ranking of MAE and MSE) in Table \ref{tab:overallmae} (PriSTI, CSDI, GRIN and BRITS) to impute all the data in AQI-36, and then use the classic spatiotemporal forecasting method Graph Wavenet \cite{wu2019graph} to make predictions on the imputed dataset. We divide the imputed dataset into training, valid and test sets as 70\%/10\%/20\%, and use the data of the past 12 time steps to predict the next 12 time steps. We use the MAE and RMSE (the square root of MSE) for evaluation. The prediction results is shown in Table \ref{tab:prediction}. Ori. represents the raw data without imputation. 
The results in Table \ref{tab:prediction} indicate that the prediction performance is affected by the data integrity, and the prediction performance on the data imputed by different methods also conforms to the imputation performance of these methods in Table \ref{tab:overallmae}. This demonstrates that our method can also help the downstream tasks after imputation.


\subsubsection{Sensitivity analysis (RQ2)}

It is obvious that the performance of model imputation is greatly affected by the distribution and quantity of observed data. For spatiotemporal imputation, sparse and continuously missing data is not conducive to the model learning the spatiotemporal correlation.
To test the imputation performance of PriSTI when the data is extremely sparse, we evaluate the imputation ability of PriSTI in the case of 10\%-90\% missing rate compared with the three baselines with the best imputation performance (BRITS, GRIN and CSDI). 
We evaluate in the block-missing and point-missing patterns of METR-LA, respectively. To simulate the sparser data in different missing patterns, for the block missing pattern, we increase the probability of consecutive missing whose lengths in the range $[12, 48]$; for the point missing pattern, we randomly drop the observed values according to the missing rate. 
We train one model for each method, and use the trained model to test with different missing rates for different missing patterns. For BRITS and GRIN, their models are trained on data that is randomly masked by 50\% with the corresponding missing pattern. For CSDI and PriSTI, their models are trained with the mask strategies consistent with the original experimental settings.

\begin{figure}[t]
	\centering
      \subfigure[MAE on METR-LA(Block)]{\label{fig:sens_a}
      \centering
       \includegraphics[width=0.45\columnwidth]{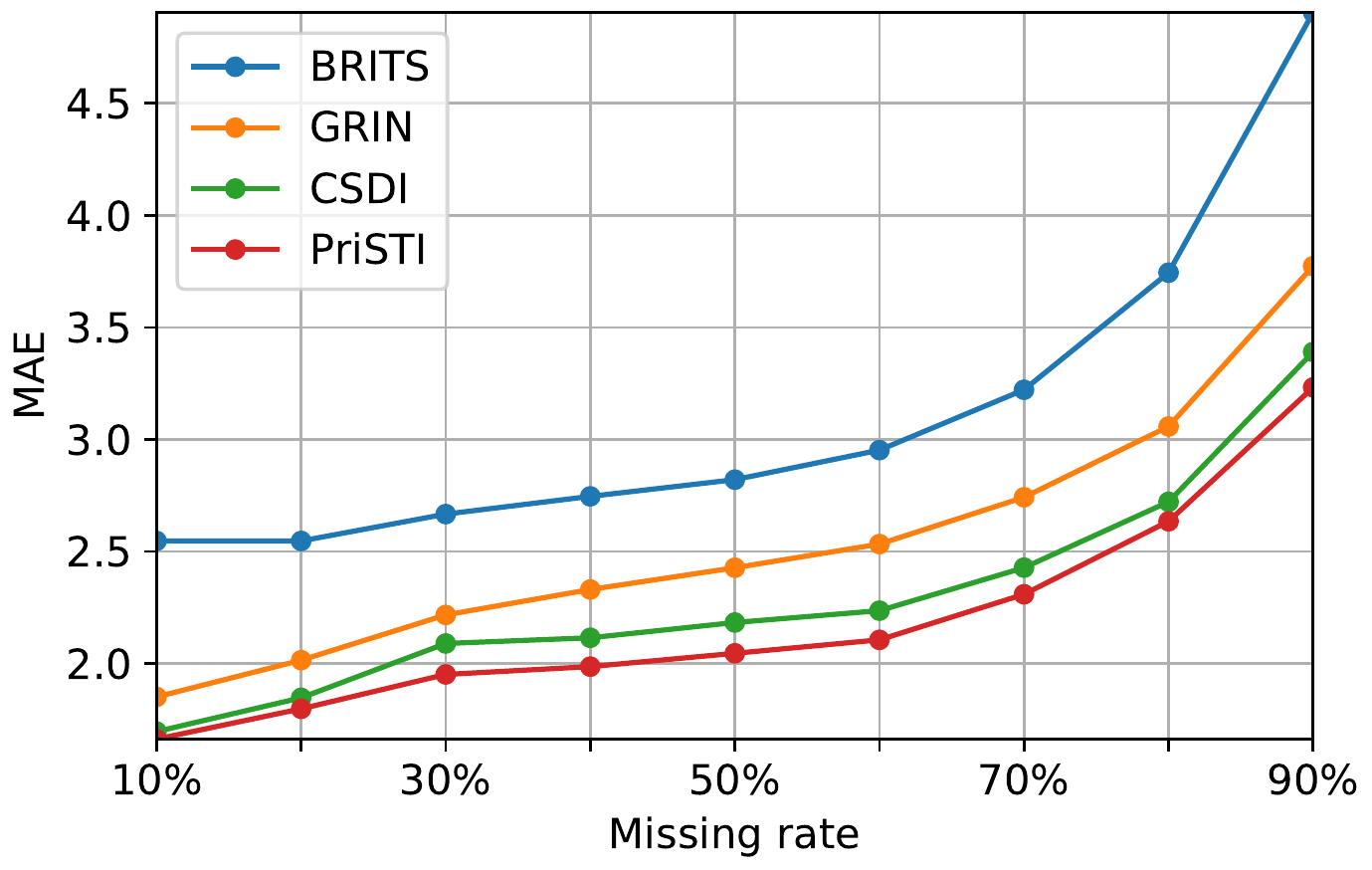}
      }
      \subfigure[MAE on METR-LA(Point)]{\label{fig:sens_b}
      \centering
       \includegraphics[width=0.45\columnwidth]{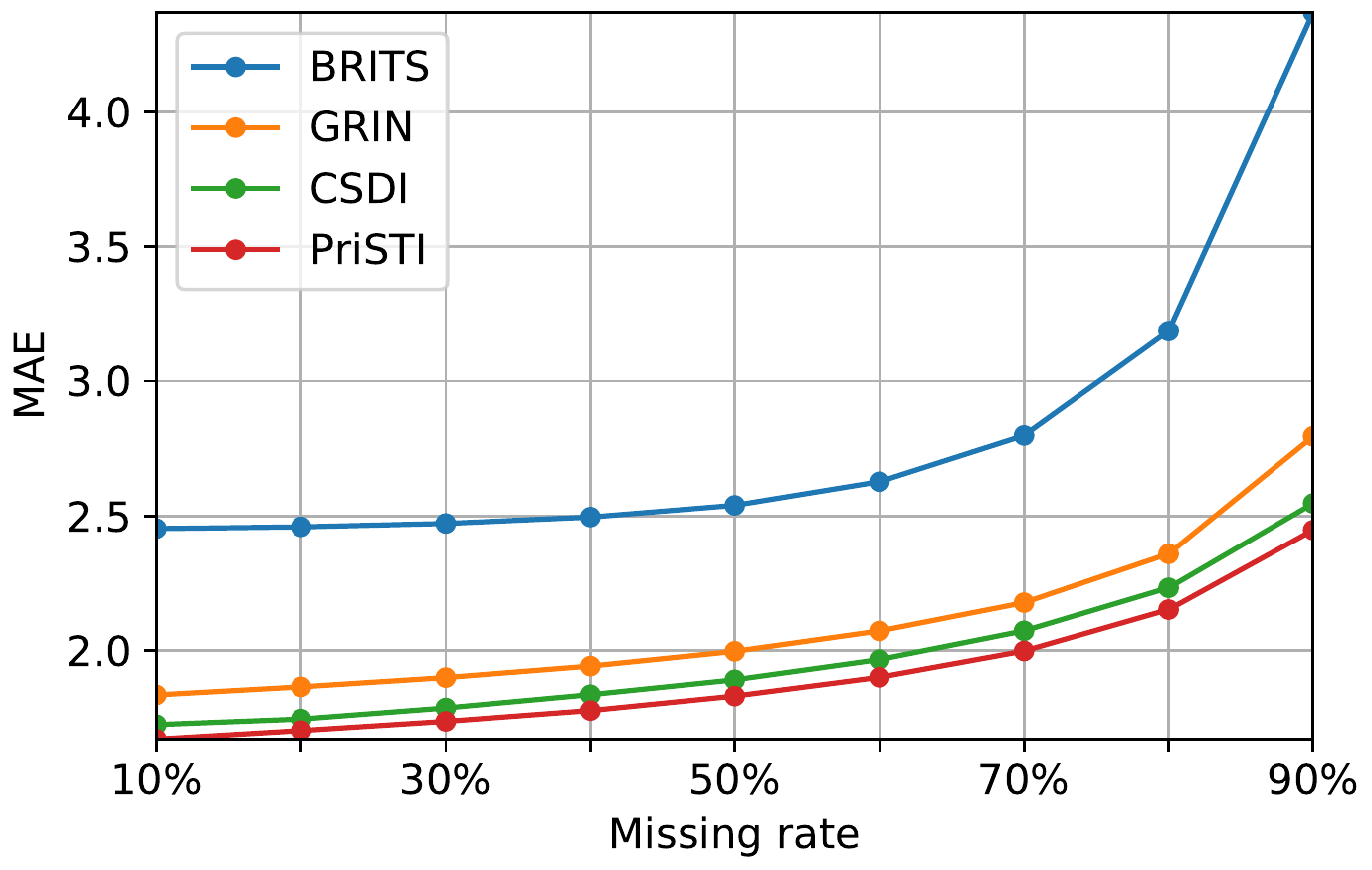}
      }
      \vspace{-2mm}
	\caption{The imputation results of different missing rates.}\label{fig:sensitivity_analysis}
 \vspace{-1mm}
\end{figure}

The MAE of each method under different missing rates are shown in Figure \ref{fig:sensitivity_analysis}. When the missing rate of METR-LA reaches 90\%, PriSTI improves the MAE performance of other methods by 4.67\%-34.11\% in block-missing pattern and 3.89\%-43.99\% in point-missing pattern.
The result indicates that our method still has better imputation performance than other baselines at high missing rate, and has a greater improvement than other methods when data is sparser.
We believe that this is due to the interpolated conditional information we construct retains the spatiotemporal dependencies that are more in line with the real distribution than the added Gaussian noise as much as possible when the data is highly sparse.

\subsubsection{Ablation study (RQ3 and RQ4)}

We design the ablation study to evaluate the effectiveness of the conditional feature extraction module and noise estimation module. We compare our method with the following variants: 
\begin{itemize}
    \item  \textit{mix-STI}: the input of noise estimation module is the concatenation of the observed values $X$ and sampled noise $\widetilde{X}^T$, and the interpolated conditional information $\mathcal{X}$ and conditional feature extraction module are not used.
    \item  \textit{w/o CF}: remove conditional features in Equation (\ref{eq:cross_att}) and  (\ref{eq:cross_att_spa}), i.e., the conditional feature extraction module is not used, and all the $Q$, $K$, and $V$ are the concatenation of interpolated conditional information $\mathcal{X}$ and sampled noise $\widetilde{X}^T$ when calculating the attention weights. 
    \item  \textit{w/o spa}: remove the spatial dependencies learning module $\gamma_\mathcal{S}$ in Equation (\ref{eq:gps_deep}). 
    \item  \textit{w/o tem}: remove the temporal dependencies learning module $\gamma_\mathcal{T}$ in Equation (\ref{eq:gps_deep}).
    \item  \textit{w/o MPNN}: remove the component of message passing neural network $\varphi_{\text{MP}}$ in the spatial dependencies learning module $\gamma_\mathcal{S}$. 
    \item  \textit{w/o Attn}: remove the component of spatial global attention $\varphi_{\text{SA}}$ in the spatial dependencies learning $\gamma_\mathcal{S}$.  
\end{itemize}

\begin{table}[t]
  \centering
  \caption{Ablation studies.}
  \label{tab:abl}
  \renewcommand{\arraystretch}{1}
  \resizebox{0.9\columnwidth}{!}{
    \begin{tabular}{cccc}
    \toprule
    \multirow{2}{*}{Method}&
    AQI-36 & \multicolumn{2}{c}{METR-LA}\cr
   \cmidrule(lr){2-2} \cmidrule(lr){3-4} 
    & {Simulated failure}& {Block-missing}& {Point-missing}\cr
    \midrule
    mix-STI & 9.83$\pm$0.04 & 1.93$\pm$0.00 & 1.74$\pm$0.00 \cr
    w/o CF & 9.28$\pm$0.05 & 1.95$\pm$0.00 & 1.75$\pm$0.00 \cr
    \midrule
    w/o spa & 10.07$\pm$0.04 & 3.51$\pm$0.01 & 2.23$\pm$0.00 \cr
    w/o tem & 10.95$\pm$0.08 & 2.43$\pm$0.00 & 1.92$\pm$0.00 \cr
    w/o MPNN & 9.10$\pm$0.10 & 1.92$\pm$0.00 & 1.75$\pm$0.00 \cr
    w/o Attn & 9.15$\pm$0.08 & 1.91$\pm$0.00 & 1.74$\pm$0.00 \cr
    \midrule
    PriSTI & \textbf{9.03$\pm$0.07}  & \textbf{1.86$\pm$0.00}  & \textbf{1.72$\pm$0.00}\cr
    \bottomrule
    \end{tabular}}
\end{table}

The variants \textit{mix-STI} and \textit{w/o CF} are used to evaluate the effectiveness of the construction and utilization of the conditional information, where \textit{w/o CF} utilizes the interpolated information $\mathcal{X}$ while \textit{mix-STI} does not.
The remaining variants are used to evaluate the spatiotemporal dependencies learning of PriSTI. \textit{w/o spa} and \textit{w/o tem} are used to prove the necessity of learning temporal and spatial dependencies in spatiotemporal imputation, and \textit{w/o MPNN} and \textit{w/o Attn} are used to evaluate the effectiveness of spatial global correlation and geographic dependency. 
Since the spatiotemporal dependencies and missing patterns of the two traffic datasets are similar, we perform ablation study on datasets AQI-36 and METR-LA, and their MAE results are shown in Table \ref{tab:abl}, from which we have the following observations: 
(1) According to the results of \textit{mix-STI}, the enhanced conditional information and the extraction of conditional feature is effective for spatiotemporal imputation. We believe that the interpolated conditional information is effective for the continuous missing, such as the simulated failure in AQI-36 and the block-missing in traffic datasets. 
And the results of \textit{w/o CF} indicate that the construction and utilization of conditional feature improve the imputation performance for diffusion model, which can demonstrate that the conditional feature extraction module and the attention weight calculation in the noise estimation module is beneficial for the spatiotemporal imputation of PriSTI, since they model the spatiotemporal global correlation with the less noisy information.
2) The results of \textit{w/o spa} and \textit{w/o tem} indicate that both temporal and spatial dependencies are necessary for the imputation. This demonstrate that our proposed noise estimation module captures the spatiotemporal dependencies based on the conditional feature, which will also be validated qualitatively in Section \ref{sec:case_study}.
3) From the results of \textit{w/o MPNN} and \textit{w/o Attn}, the components of the spatial global attention and message passing neural network have similar effects on imputation results, but using only one of them is not as effective as using both, which indicates that spatial global correlation and geographic information are both necessary for the spatial dependencies.
Whether the lack of geographic information as input or the lack of captured implicit spatial correlations, the imputation performance of the model would be affected.
We believe that the combination of explicit spatial relationships and implicit spatial correlations can extract useful spatial dependencies in real-world datasets.

\begin{figure}[t]
	\centering
      \subfigure[AQI-36]{\label{fig:visual_aqi}
      \centering
       \includegraphics[width=0.47\columnwidth]{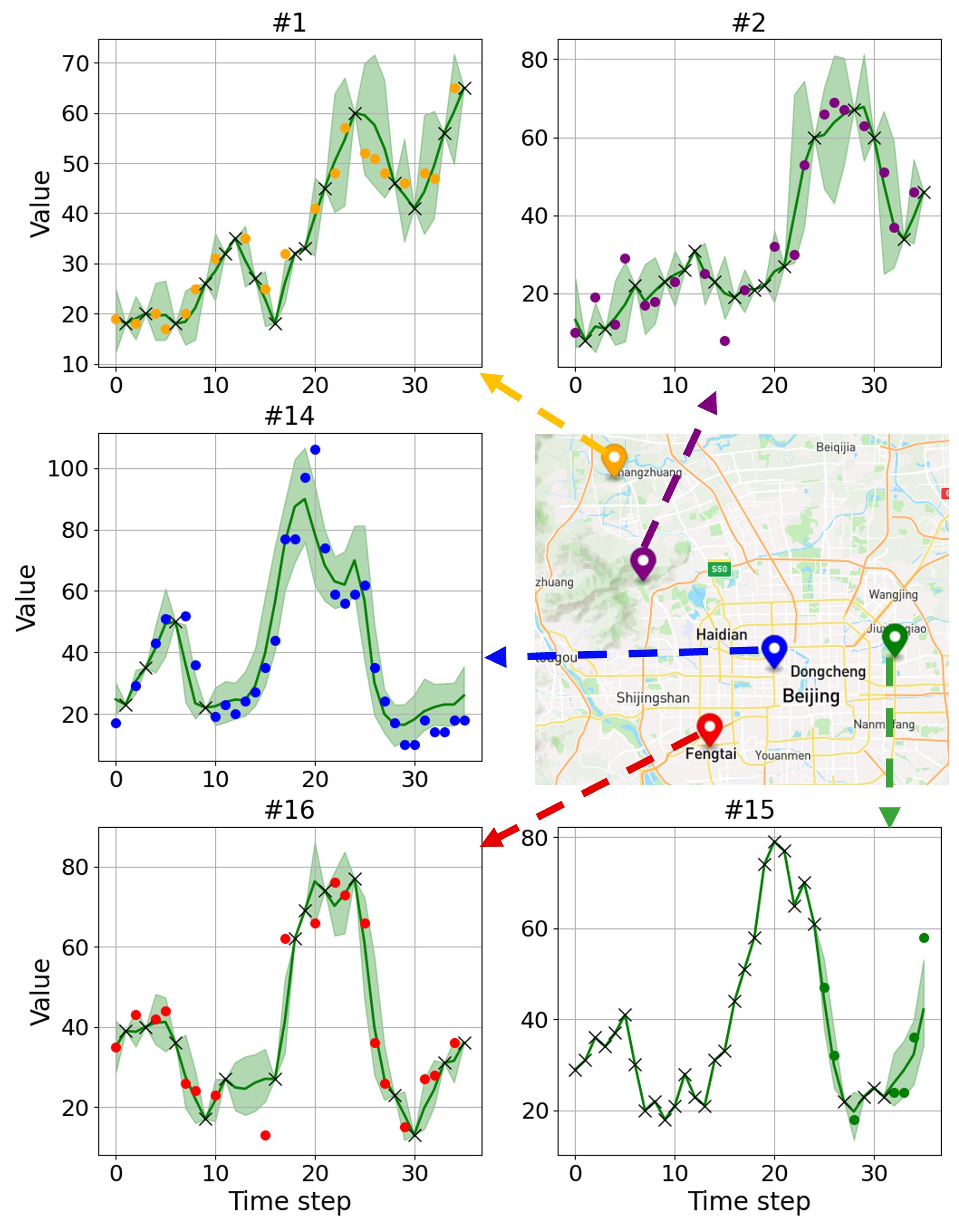}
      }
      \subfigure[METR-LA (Block)]{\label{fig:visual_la_block}
      \centering
       \includegraphics[width=0.45\columnwidth]{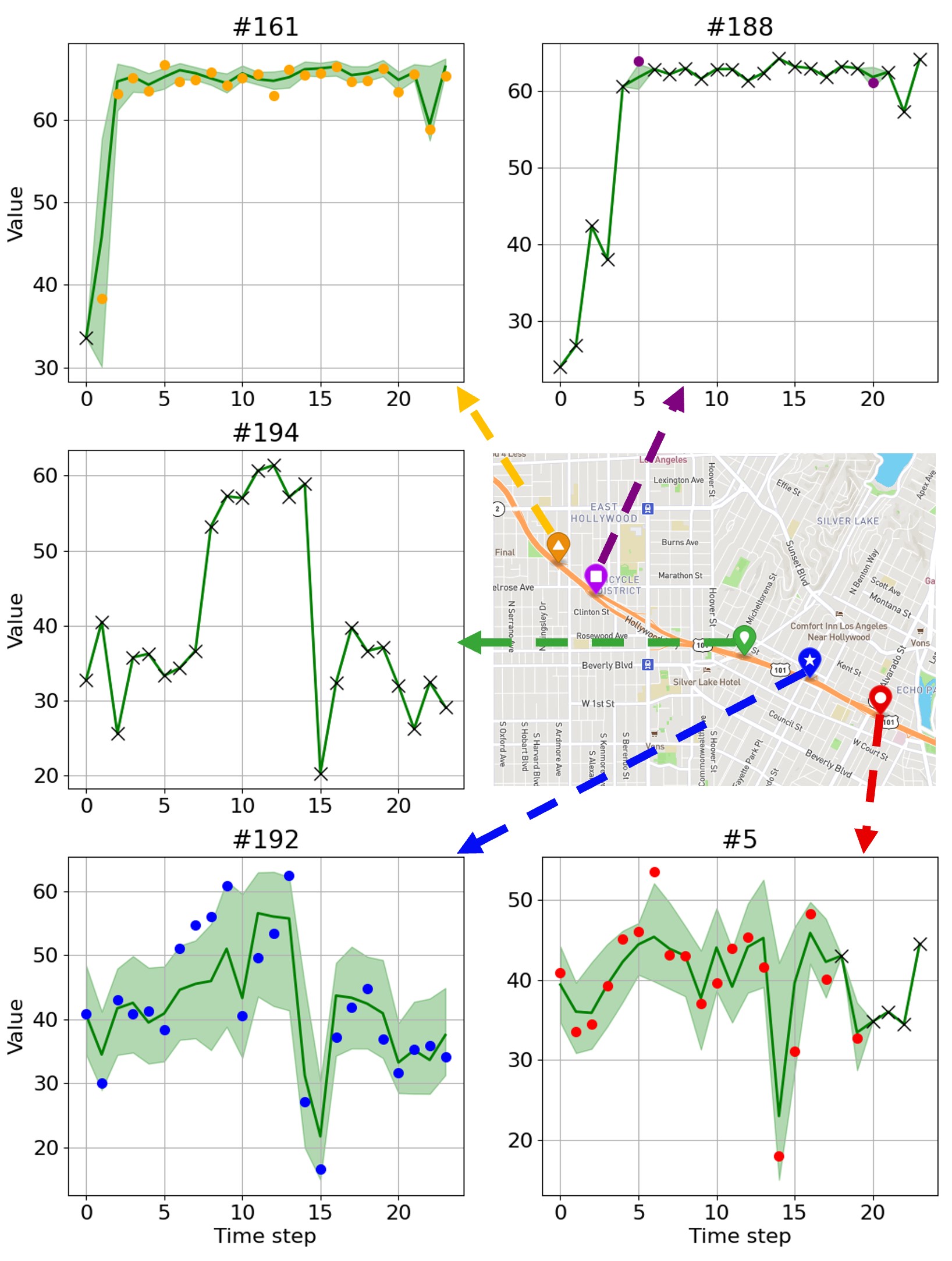}
      }      
	\caption{The visualization of the probabilistic imputation in AQI-36 and block-missing pattern of METR-LA. Each subfigure represents a sensor, and the time windows of all sensors are aligned. The black crosses represent observations, and dots of various colors represent the ground truth of missing values. The solid green line is the deterministic imputation result, and the green shadow represents the quantile between 0.05 to 0.95.}\label{fig:visual_la_block}
    \vspace{-3mm}
\end{figure}

\subsubsection{Case study (RQ4)}\label{sec:case_study}

We plot the imputation result at the same time of some nodes in AQI-36 and the block-missing pattern of METR-LA to qualitatively analyze the spatiotemporal imputation performance of our method, as shown in Figure \ref{fig:visual_la_block}. 
Each of the subfigure represents a sensor, the black cross represents the observed value, and the dots of other colors represent the ground truth of the part to be imputed. The green area is the part between the 0.05 and 0.95 quantiles of the estimated probability distribution, and the green solid line is the median, that is, the deterministic imputation result.

We select 5 sensors in AQI-36 and METR-LA respectively, and display their geographic location on the map. Taking METR-LA as the example, it can be observed from Figure \ref{fig:visual_la_block} that sensor 188 and 194 have almost no missing values in the current time window, while their surrounding sensors have continuous missing values, and sensor 192 even has no observed values, which means its temporal information is totally unavailable. 
However, the distribution of our generated samples still covers most observations, and the  imputation results conform to the time trend of different nodes. 
This indicates that on the one hand, our method can capture the temporal dependencies by the given observations for imputation, and on the other hand, when the given observations are limited, our method can utilize spatial dependencies to impute according to the geographical proximity or nodes with similar temporal pattern. For example, for the traffic system, the time series corresponding to sensors with close geographical distance are more likely to have similar trends.

\begin{figure}[t]
	\center
	\includegraphics[width=0.48\textwidth]{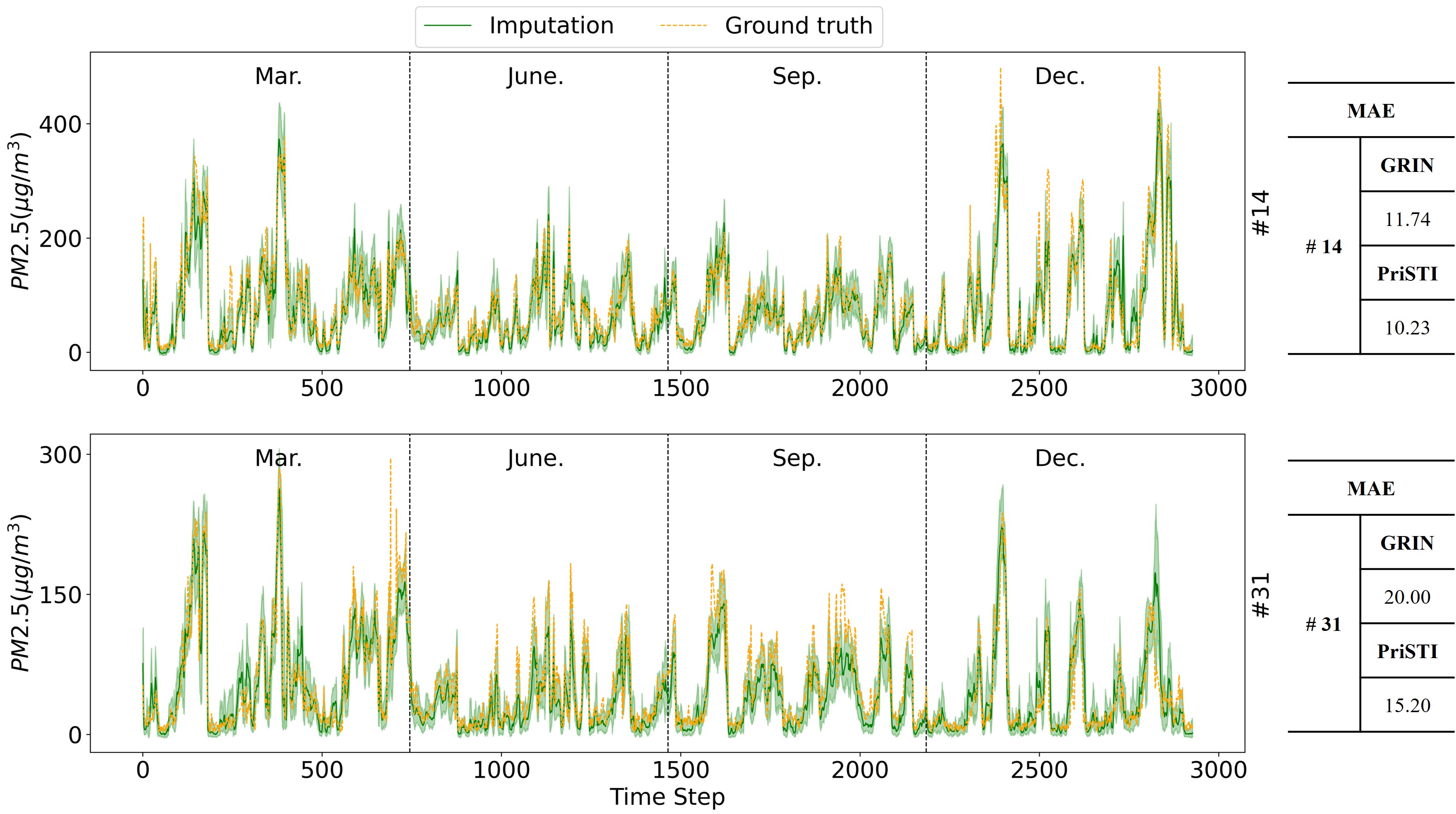}
	\caption{The imputation for unobserved sensors in AQI-36. The orange dotted line represents the ground truth, the green solid line represents the deterministic imputation result, and the green shadow represents the quantile between 0.05 to 0.95.}\label{fig:mask_sensor}
    \vspace{-3mm}
\end{figure}

\subsubsection{Imputation for sensor failure (RQ5)}

We impute the spatiotemporal data in a more extreme case: some sensors fail completely, i.e., they cannot provide any observations. The available information is only their location, so we can only impute by the observations from other sensors. 
This task is often studied in the research related to Kriging \cite{stein1999interpolation}, which requires the model to reconstruct a time series for a given location based on geographic location and observations from other sensors.

We perform the experiment of sensor failure on AQI-36.  According to \cite{cini2021filling}, we select the air quality monitoring station with the highest (station 14) and lowest (station 31) connectivity. Based on the original experimental settings unchanged, all observations of these two nodes are masked during the training process. The results of the imputation for unobserved sensors are shown in Figure \ref{fig:mask_sensor}, where the orange dotted line is the ground truth, the green solid line is the median of the generated samples, and the green shadow is the quantiles between 0.05 and 0.95. 
We use MAE to quantitatively evaluate the results, the MAE of station 14 is 10.23, and the MAE of station 31 is 15.20. 
Since among all baselines only GRIN can impute by geographic information, the MAE compared to the same experiments in GRIN is also shown in Figure \ref{fig:mask_sensor}, and PriSTI has a better imputation performance on the unobserved nodes.
This demonstrates the effectiveness of PriSTI in exploiting spatial relationship for imputation. 
Assuming that the detected spatiotemporal data are sufficiently dependent on spatial geographic location, our proposed method may be capable of reconstructing the time series for a given location within the study area, even if no sensors are deployed at that location.

\subsubsection{Hyperparameter analysis and time costs}

\begin{figure}[t]
	\centering
      \subfigure[Channel size $d$]{\label{fig:hp_a}
      \centering
       \includegraphics[width=0.28\columnwidth]{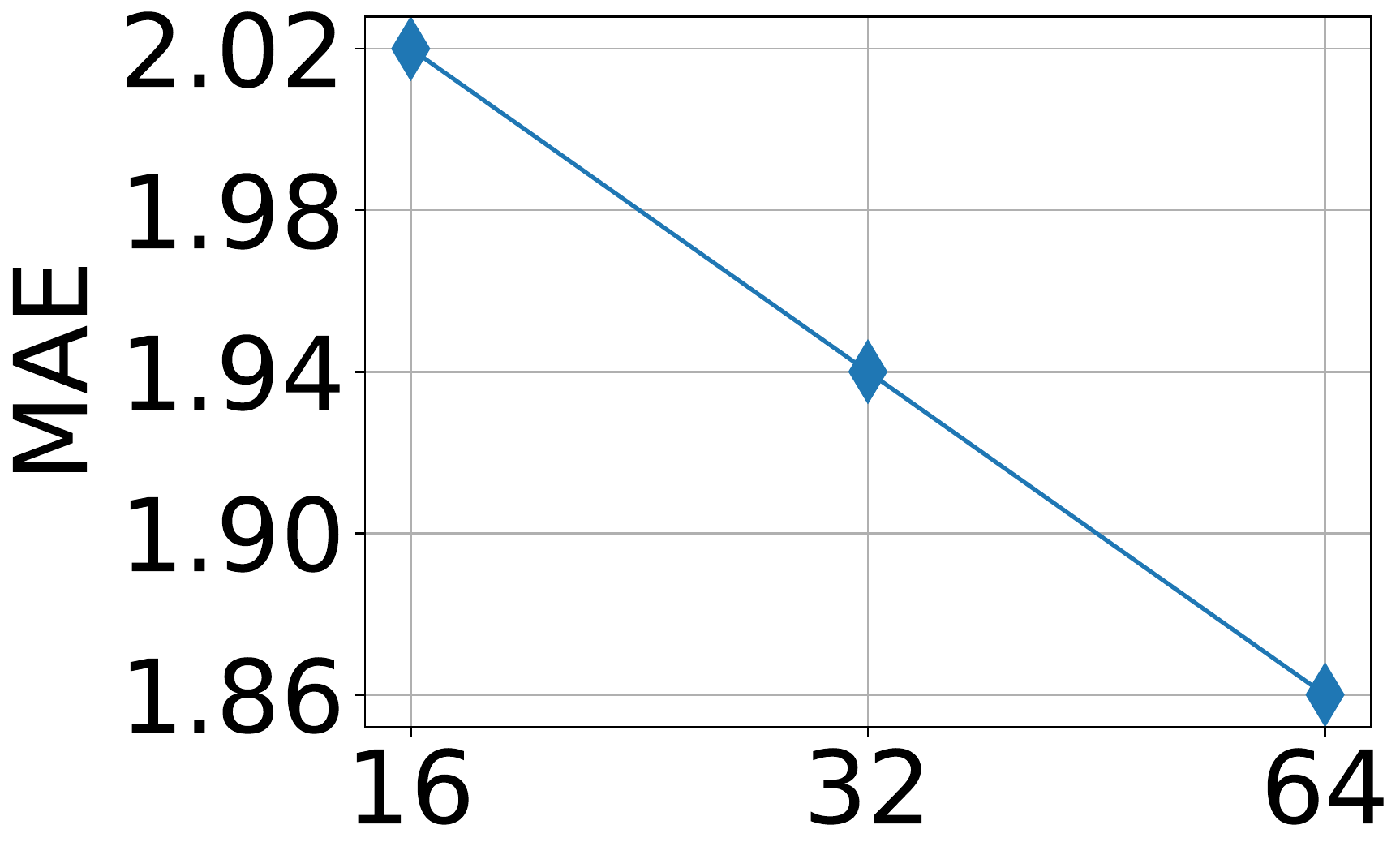}
      }
      \subfigure[Maximum noise level $\beta_T$]{\label{fig:hp_b}
      \centering
       \includegraphics[width=0.28\columnwidth]{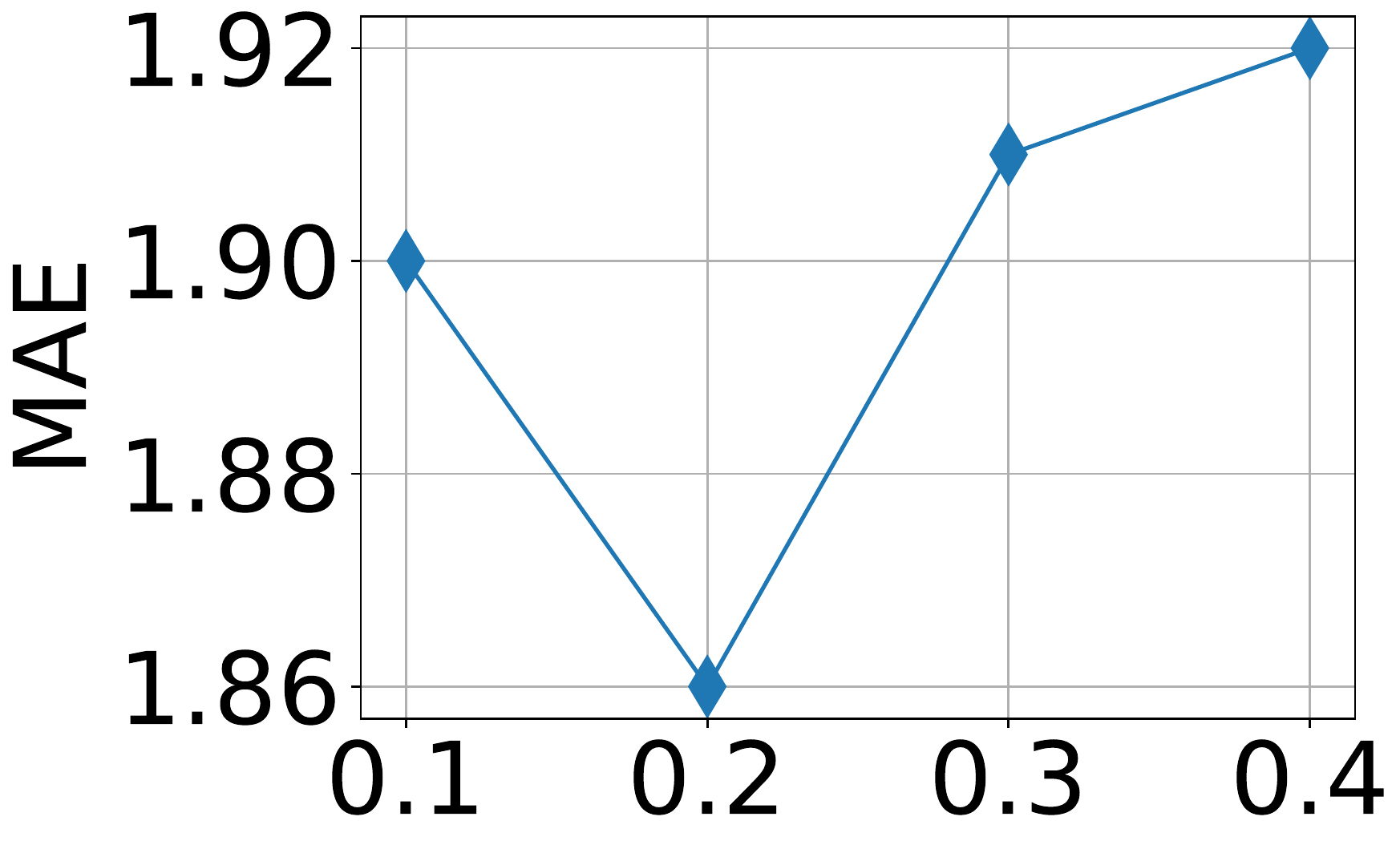}
      }
      \subfigure[Number of virtual nodes $k$]{\label{fig:hp_c}
      \centering
       \includegraphics[width=0.28\columnwidth]{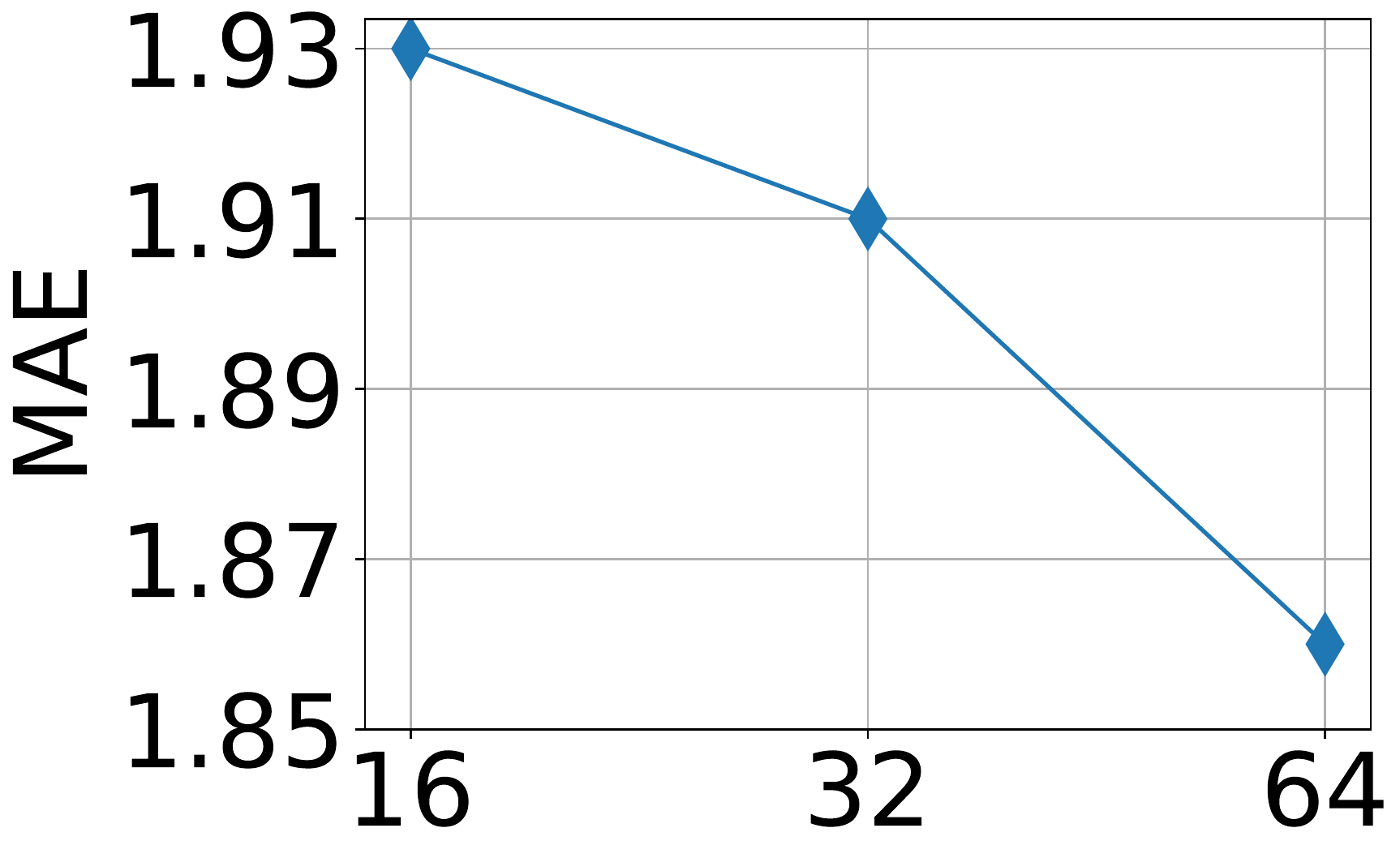}
      }
      \vspace{-2mm}
	\caption{The sensitivity study of key parameters.}\label{fig:sensitivity_parameter}
 \vspace{-3mm}
\end{figure}

We conduct analysis and experiments on some key hyperparameters in PriSTI to illustrate the setting of hyperparameters and the sensitivity of the model to these parameters. Taking METR-LA as example, we analyze the three key hyperparameters: the channel size of hidden state $d$, maximum noise level $\beta_T$ and number of virtual nodes $k$, as shown in Figure \ref{fig:sensitivity_parameter}.
Among them, $d$ and $k$ affect the amount of information learned by the model. $\beta_T$ affects the level of sampled noise, and too large or too too small value is not conducive to the learning of noise prediction model.
According to the results in Figure \ref{fig:sensitivity_parameter}, we set 0.2 to $\beta_T$ which is the optimal value. For $d$ and $k$, although the performance is better with larger value, we set both $d$ and $k$ to 64 considering the efficiency.

\begin{figure}[t]
	\centering
      \subfigure[Training time]{\label{fig:eff_a}
      \centering
       \includegraphics[width=0.4\columnwidth]{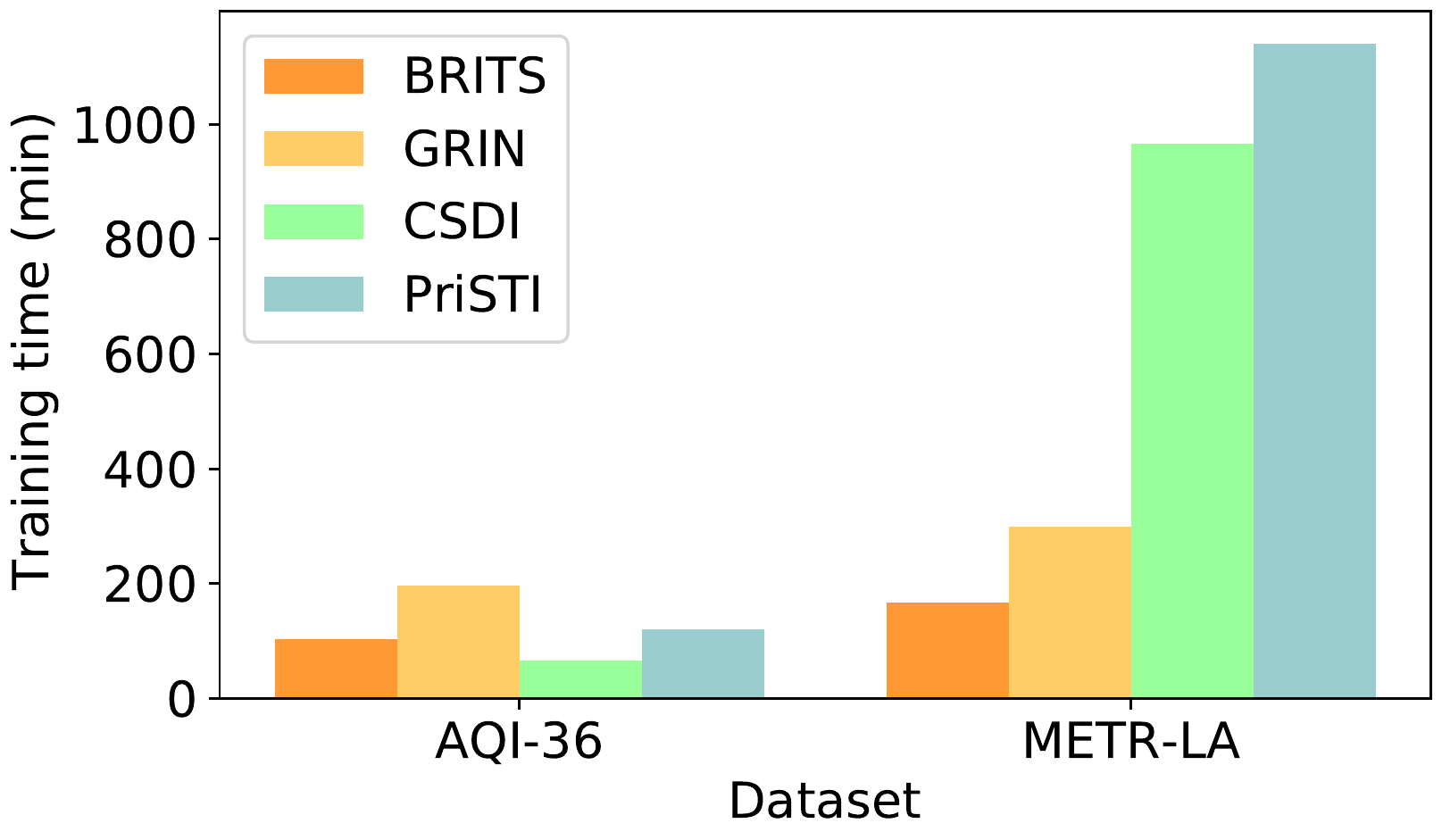}
      }
      \subfigure[Inference time]{\label{fig:eff_b}
      \centering
       \includegraphics[width=0.4\columnwidth]{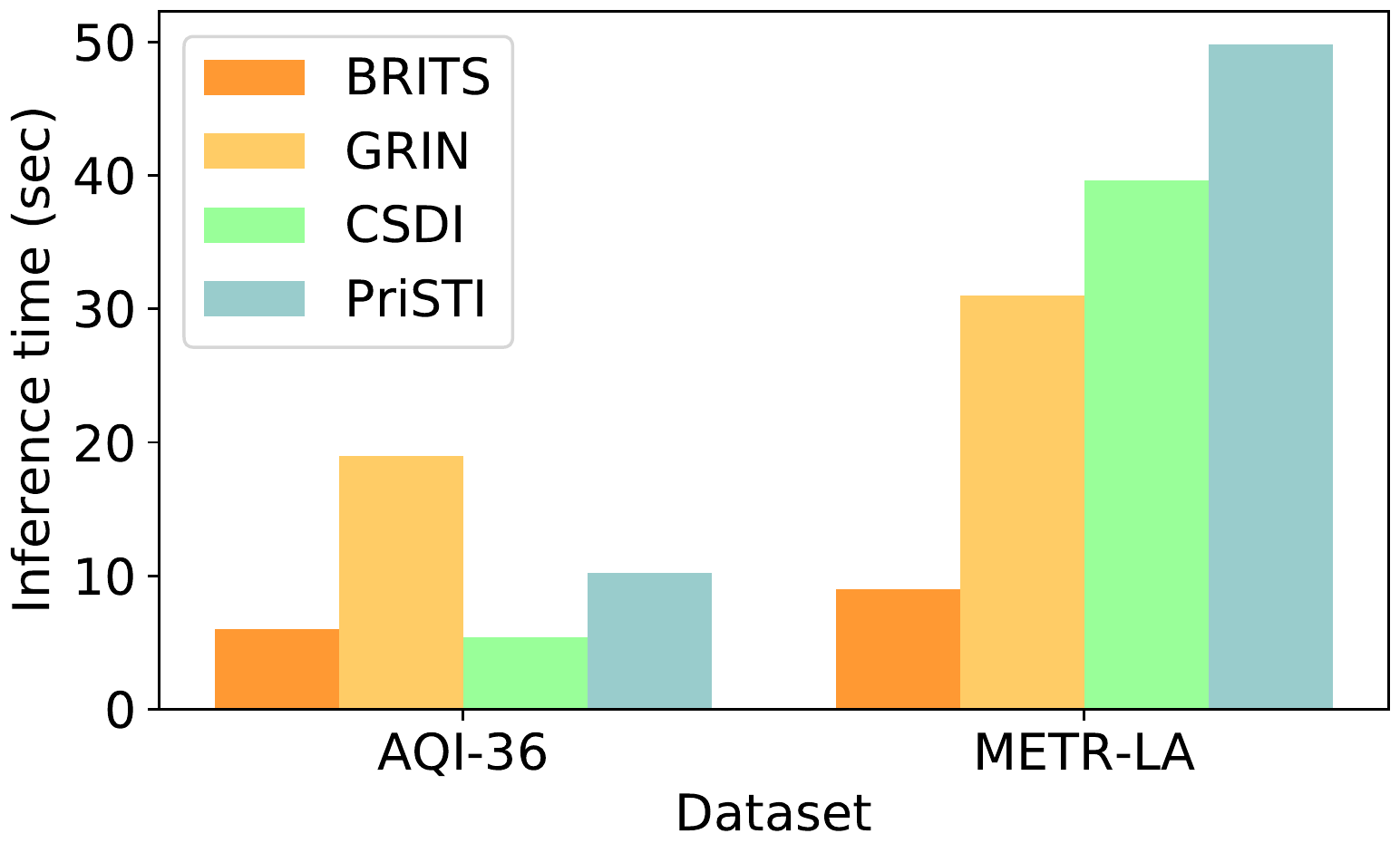}
      }      
      \vspace{-2mm}
	\caption{The time costs of PriSTI and other baselines.}\label{fig:efficiency}
 \vspace{-3mm}
\end{figure}

In addition, we select several deep learning methods with the higher imputation performance ranking to compare their efficiency with PriSTI on dataset AQI-36 and METR-LA. The total training time and inference time of different methods are shown in Figure \ref{fig:efficiency}. 
The experiments are conducted on AMD EPYC 7371 CPU, NVIDIA RTX 3090. 
It can be seen that the efficiency gap between methods on datasets with less nodes (AQI-36) is not large, but the training time cost of generative methods CSDI and PriSTI on datasets with more nodes (METR-LA) is higher. Due to the construction of conditional information, PriSTI has 25.7\% more training time and 17.9\% more inference time on METR-LA than CSDI.

%% file: 5_Related_Work.tex
\section{Related Work}\label{sec:related_work}


Since the spatiotemporal data can be imputed along temporal or spatial dimension, there are a large amount of literature for missing value imputation in spatiotemporal data. 
For the time series imputation, the early studies imputed missing values by statistical methods such as local interpolation \cite{kreindler2006effects, acuna2004treatment}, which reconstructs the missing value by fitting a smooth curve to the observations. 
Some methods also impute missing values based on the historical time series through EM algorithm \cite{shumway1982approach, nelwamondo2007missing} or the combination of ARIMA and Kalman Filter \cite{harvey1990forecasting, ansley1984estimation}.
There are also some early studies filling in missing values through spatial relationship or neighboring sequence, such as KNN \cite{trevor2009elements, beretta2016nearest} and Kriging \cite{stein1999interpolation}.
In addition, the low-rank matrix factorization \cite{salakhutdinov2008bayesian, yu2016temporal, chen2019missing, chen2021bayesian} is also a feasible approach for spatiotemporal imputation, which exploits the intrinsic spatial and temporal patterns based on the prior knowledge.
For instance, TRMF \cite{yu2016temporal} incorporates the structure of temporal dependencies into a temporal regularized matrix factorization framework.
BATF \cite{chen2019missing} incorporates domain knowledge from transportation systems into an augmented tensor factorization model for traffic data modeling.

In recent years, there have been many studies on spatiotemporal data imputation through deep learning methods \cite{liu2019naomi, ma2019cdsa}. 
Most deep learning imputation methods focus on the multivariate time series and use RNN as the core to model temporal relationships \cite{che2018recurrent, yoon2018estimating, cao2018brits, cini2021filling}.  
The RNN-based approach for imputation is first proposed by GRU-D \cite{che2018recurrent} and is widely used in deep autoregressive imputation methods. Among the RNN-based methods, BRITS \cite{cao2018brits} imputes the missing values on the hidden state through a bidirectional RNN and considers the correlation between features. 
GRIN \cite{cini2021filling} introduces graph neural networks based on BRITS to exploit the inductive bias of historical spatial patterns for imputation. 
In addition to directly using RNN to estimate the hidden state of missing parts, there are also a number of methods using GAN to generate missing data \cite{luo2018multivariate, yoon2018gain, miao2021generative}. 
For instance, GAIN \cite{yoon2018gain} imputes data conditioned on observation values by the generator, and utilizes the discriminator to distinguish the observed and imputed part.
SSGAN \cite{miao2021generative} proposes a semi-supervised GAN to drive the generator to estimate missing values using observed information and data labels.

However, these methods are still RNN-based autoregressive methods, which are inevitably affected by the problem of error accumulation, i.e., the current missing value is imputed by the inaccurate historical estimated values in a sequential manner. 
To address this problem, Liu et al. \cite{liu2019naomi} proposes NAOMI, developing a non-autoregressive decoder that recursively updates the hidden state, and using generative adversarial training to impute. 
Fortuin et al. \cite{fortuin2020gp} propose a multivariate time series imputation method utilizing the VAE architecture with a Gaussian process prior in the latent space to capture temporal dynamics.
Some other works capture spatiotemporal dependencies through the attention mechanism \cite{ma2019cdsa, shukla2021multi, du2022saits}, which not only consider the temporal dependencies but also exploit the geographic locations \cite{ma2019cdsa} and correlations between different time series \cite{du2022saits}.

Recently, a diffusion model-based generative imputation framework CSDI \cite{tashiro2021csdi} shows the performance advantages of deep generative models in multivariate time series imputation tasks.
The Diffusion Probabilistic Models (DPM) \cite{sohl2015deep, ho2020denoising, song2020score}, as deep generative models, have achieved great performance than other generative methods in several fields such as image synthesis \cite{rombach2022high, ho2020denoising}, audio generation \cite{kong2020diffwave, goel2022s}, and graph generation \cite{huang2022graphgdp, huang2023conditional}.  
In terms of imputation tasks, there are existing methods for 3D point cloud completion \cite{lyu2021conditional} and multivariate time series imputation \cite{tashiro2021csdi} through conditional DPM.  
CSDI imputes the missing data through score-based diffusion models conditioned on observed data, exploiting temporal and feature correlations by a two dimensional attention mechanism.
However, CSDI takes the concatenation of observed values and noisy information as the input when training, increasing the difficulty of the attention mechanism's learning. 
Different from existing diffusion model-based imputation methods, our proposed method construct the prior and imputes spatiotemporal data based on the extracted conditional feature and geographic information.


%% file: 6_Conclusion.tex
\section{Conclusion}\label{sec:conclusion}

We propose PriSTI, a conditional diffusion framework for spatiotemporal imputation, which imputes missing values with help of the extracted conditional feature to calculate temporal and spatial global correlations. 
Our proposed framework captures spatiotemporal dependencies by comprehensively considering spatiotemporal global correlation and geographic dependency. 
PriSTI achieves more accurate imputation results than state-of-the-art baselines in various missing patterns of spatiotemporal data in different fields, and also handles the case of high missing rates and sensor failure.
In future work, we will consider improving the scalability and computation efficiency of existing frameworks on larger scale spatiotemporal datasets, and how to impute by longer temporal dependencies with refined conditional information.